\documentclass[nohyperref]{article}

\usepackage{microtype}
\usepackage{graphicx}
\usepackage{subfigure}
\usepackage{booktabs} %

\usepackage{hyperref}

\usepackage[accepted]{icml2022}

\usepackage{amsmath}
\usepackage{amssymb}
\usepackage{mathtools}
\usepackage{amsthm}

\usepackage{graphicx}

\newcommand{\EE}{\mathbb{E}}
\usepackage{stmaryrd}
\usepackage{array}
\usepackage{tabularx}
\usepackage{comment}
\everypar{\looseness=-1}

\usepackage[capitalize,noabbrev]{cleveref}

\theoremstyle{plain}
\newtheorem{theorem}{Theorem}[section]

\newtheorem{lemma}[theorem]{Lemma}

\theoremstyle{definition}

\theoremstyle{remark}

\usepackage[textsize=tiny]{todonotes}

\icmltitlerunning{Large Batch Experience Replay}

\begin{document}

\twocolumn[
\icmltitle{Large Batch Experience Replay}

\icmlsetsymbol{equal}{*}

\begin{icmlauthorlist}
\icmlauthor{Thibault Lahire}{sch}
\icmlauthor{Matthieu Geist}{comp}
\icmlauthor{Emmanuel Rachelson}{sch}
\end{icmlauthorlist}

\icmlaffiliation{sch}{ISAE-SUPAERO, Université de Toulouse, France}
\icmlaffiliation{comp}{Google Research, Brain Team}

\icmlcorrespondingauthor{Thibault Lahire}{thibaultlahire.research@gmail.com}

\icmlkeywords{Reinforcement Learning, Importance Sampling, Stochastic Gradient Descent, Experience Replay, ICML}

\vskip 0.3in
]

\printAffiliationsAndNotice{} 

\begin{abstract}
Several algorithms have been proposed to sample non-uniformly the replay buffer of deep Reinforcement Learning (RL) agents to speed-up learning, but very few theoretical foundations of these sampling schemes have been provided. 
Among others, Prioritized Experience Replay appears as a hyperparameter sensitive heuristic, even though it can provide good performance. 
In this work, we cast the replay buffer sampling problem as an importance sampling one for estimating the gradient. 
This allows deriving the theoretically optimal sampling distribution, yielding the best theoretical convergence speed. %
Elaborating on the knowledge of the ideal sampling scheme, we exhibit new theoretical foundations of Prioritized Experience Replay. 
The optimal sampling distribution being intractable, we make several approximations providing good results in practice and introduce, among others, LaBER (Large Batch Experience Replay), an easy-to-code and efficient method for sampling the replay buffer. 
LaBER, which can be combined with Deep Q-Networks, distributional RL agents or actor-critic methods, yields improved performance over a diverse range of Atari games and PyBullet environments, compared to the base agent it is implemented on and to other prioritization schemes.
\end{abstract}

\section{Introduction}

\looseness=-1
In deep Reinforcement Learning \citep[RL]{sutton2018reinforcement}, neural network policies and value functions can be learnt thanks to stochastic gradient descent algorithms \citep[SGD]{robbins1951stochastic} sampling an experience replay memory \citep{lin1992self}. 
This replay memory, or replay buffer, stores the transitions encountered along the interaction with the environment. 
SGD-based algorithms exploit such buffers to learn relevant functions, such as the Q-function in the case of Deep Q-Networks \citep[DQN]{mnih2015human}, the return distribution for distributional approaches \citep{bellemare2017distributional}, or an actor and a critic \citep{lillicrap2016continuous, Haarnoja2018SAC} in the case of continuous state-action space problems. 
Most Deep RL algorithms boil down to a sequence of SGD-based, supervised learning problems.
Ideally, one would like to minimize the corresponding loss functions with as few gradient steps as possible. 

Prioritized Experience Replay \citep[PER]{Schaul2015prioritized} has been introduced for the DQN algorithm as a heuristic accelerating the learning process, drawing inspiration from Prioritized Sweeping \citep{Moore1993PrioritizedSweeping}.
Each transition in the replay buffer is assigned a priority that is proportional to the temporal difference (TD) error, and gradients are estimated by sampling according to these priorities. 
Later, PER was combined with other DQN improvements, including distributional RL \citep{bellemare2017distributional,Hessel2018Rainbow}, as well as applied to actor-critic methods \citep{wang2019boosting}. 
However the reason why prioritizing on TD errors can provide better performance remains theoretically unexplained.
For distributional value functions, using the loss value as a priority, as suggested by \citet{Hessel2018Rainbow}, also lacks foundations. 
Lastly, in actor-critic methods, when two critics are used \citep{fujimoto2018addressing,Haarnoja2018SAC}, two TD errors are available and the appropriate way to draw mini-batches has, again, not been established clearly. 
In this work, we provide theoretical foundations for the concept of prioritization used in RL, and derive the associated algorithms, shedding light on existing approaches on the way.

SGD relies on the fact that the loss minimized when learning a neural network is an integral quantity over a certain distribution.
Hence, the gradient of this loss is also such an integral, which can be approximated via Monte Carlo estimation with a finite number of samples.
The variance of this Monte Carlo estimate can be reduced using importance sampling \citep{rubinstein2016simulation}, yielding better convergence speed.
Consequently, non-uniform sampling of mini-batches for gradient estimation can be cast as an importance sampling problem over the replay buffer, such as in the work of~\citet{Needell2014SGD}. %
The Supervised Learning literature provides links between sampling, variance of the stochastic gradient estimate, and convergence speed of the SGD algorithm \citep{wang2017accelerating}. 
The smaller the variance of the stochastic gradient estimate, the faster the convergence. 
This is particularly appealing in the context of Approximate Dynamic Programming (ADP), which encompasses the vast majority of (Deep) RL algorithms (e.g. DQN, DDPG \citep{lillicrap2016continuous}, SAC \citep{Haarnoja2018SAC} and their variations). 
ADP is very sensitive to approximation errors and only a few noisy gradient steps are taken at each Dynamic Programming iteration in modern Deep RL algorithms.
The optimal sampling distribution, proportional to the per-sample gradient norms, is intractable and approximations were proposed \citep{Loshchilov2016, Idiap2018NASACE}. 
We show that PER is a special case of such approximations in the context of ADP, and propose better sampling schemes, theoretically grounded and less sensitive to hyperparameter tuning, that naturally extend to any value function representation, including distributional Q-functions or twin critics. 
In this work, we show that the main issue with PER are its outdated priorities, and introduce the Large Batch Experience Replay (LaBER) algorithm which constitutes our main algorithmic contribution.

This paper is structured as follows. 
Section 2 covers related work in prioritization for RL and importance sampling for SGD. 
Then Section 3 casts the gradient estimation problem in the light of importance sampling and proposes algorithms to mimic the intractable optimal sampling distribution.
On the way, it provides theoretical foundations to PER. 
Section 4 empirically evaluates the proposed sampling schemes, in particular the LaBER algorithm. 
We discuss each separate aspect of sampling, their explanations and perspectives. 
Section 5 summarizes and concludes.

\section{Background and Related Works}
\label{sec:related}

In the standard RL framework, one searches for the optimal control policy when interacting with a discrete-time system behaving as a Markov Decision Process \citep{puterman2014markov}. 
At time step $t$, the system is in state $s_t\in S$, and upon applying action $a_t \in A$, it transitions to a new state $s_{t+1}$, while receiving reward $r_t$. 
A policy $\pi$ is a function mapping states to distributions over actions, whose performance can be assessed through its Q-function $Q^\pi(s,a) = \mathbb{E}[\sum_t \gamma^t r_t |s_0 = s,a_0 = a, a_t \sim \pi(s_t)]$. 
The goal of RL is to find a policy $\pi^*$ that has the largest possible Q-function $Q^* = Q^{\pi^*}$.

\looseness=-1
Policy $\pi$'s Q-function obeys the fixed-point equation $Q^\pi(s,a) = \mathbb{E}_{s',r} \left[r + \gamma Q^\pi(s',\pi(s')) \right]$. 
Similarly, the optimal Q-function obeys equation $Q^*(s,a) = \mathbb{E}_{s',r} \left[r + \gamma \max_{a'} Q^*(s',a') \right]$. 
These are called the Bellman evaluation and optimality equations, respectively. 
They are often summarized by introducing the evaluation and optimality operators on functions: $Q^\pi=T^\pi Q^\pi$ and $Q^* = T^*Q^*$ respectively. 
These operators are contraction mappings, which implies that the $Q_{n+1}=T Q_n$ sequence converges to the $T$ operator's fixed point: $Q^\pi$ for $T=T^\pi$, and $Q^*$ for $T=T^*$. 
The optimality and evaluation equations are cornerstones of the vast majority of RL algorithms, as they underpin respectively the search for optimal value functions in Value Iteration methods and the evaluation of current policies in Policy Iteration and Policy Gradient methods (including actor-critic ones).
Approximation of $T Q_n$ is thus a key issue in Reinforcement Learning and gives rise to the family of Approximate Dynamic Programming (ADP) methods.

ADP is known to suffer from the approximation errors incurred by the family of Q-functions used \citep{munos2003error,munos2005error,scherrer2012approximate}. In particular, it is well-known that ADP does not converge but rather that $\|Q_n - Q^*\| \leq O(\epsilon / (1-\gamma)^2)$ for a large enough $n$, where $\epsilon$ is an upper-bound on the approximation error of $Q_n$. Approximating $Q_{n+1}$ from samples of $TQ_n$ is a Supervised Learning problem with error at most $\epsilon$ and, therefore, reducing $\epsilon$ immediately translates to better ADP algorithms.

Deep Q-Networks \citep[DQN]{mnih2015human} is the approximate Value Iteration algorithm that uses a replay buffer  of $N$ samples $(s,a,r,s')$, a deep neural network $Q_\theta$, and a few steps of gradient descent to approximate $T^* Q_n$. 
In this context, $Q_n$ is called a target network and is periodically replaced by $Q_\theta$. 
Specifically, at each training step, DQN aims to take a gradient step on the loss $\mathcal{L}_n(\theta) = \|Q_\theta - T^*Q_n \|^2$, or more generally  $\mathcal{L}_n(\theta)= \int_{S \times A} \ell(Q_\theta(s,a),T^* Q_n(s,a)) d\rho(s,a)$, with $\rho$ the state-action data distribution. 
Since this loss is an integral quantity, it can be estimated by Monte Carlo sampling, which defines the empirical loss $\frac{1}{N} \sum_{i=1}^N \ell(Q_\theta(x_i),y_i)$, with $x_i=(s_i,a_i)$ and $y_i=r_i+\gamma\max_{a'} Q_n(s'_i, a')$. 
Minimization of this empirical loss by SGD implies drawing at each step a mini-batch of $B$ transitions from the replay buffer and taking a descent step $\theta_{t+1} = \theta_t - \eta d$ in the direction of the gradient estimate $d = \frac{1}{B} \sum_{i=1}^B \nabla_{\theta} \ell(Q_\theta(x_i), y_i)$, with learning rate $\eta$. 
Similar targets and loss functions have been proposed in the context of distributional RL \citep{dabney2018distributional,dabney2018implicit} or for learning critics in policy gradient methods~\citep{fujimoto2018addressing,Haarnoja2018SAC}.

As is common in SGD algorithms, the mini-batch is drawn uniformly with replacement within the replay buffer of size $N$, yielding an unbiased estimate of $\nabla_\theta \mathcal{L}_n(\theta)$. 
The question of sampling non-uniformly the replay buffer has been raised from an empirical perspective in RL and has lead to heuristics such as PER and its variants \citep{Horgan2018ApeX, fujimoto2020equivalence}. 
In order to put more emphasis on $(s,a)$-pairs that feature a large approximation error of $T^* Q_n$, PER assigns each transition of the replay buffer a priority based on the TD error $(|\delta_i|+c)^\alpha$, where $\delta_i = Q_\theta(x_i) - y_i$ is the TD error, and with $\alpha$ and $c$ two non-negative hyperparameters. 
At each iteration, PER samples a mini-batch according to the probability distribution induced by the list of priorities, performs a gradient step and updates the priority of the selected samples. 
Even though PER lacks theoretical foundations, a large number of publications experimentally demonstrate its benefits. 
Notably, the ablation study of \citet{Hessel2018Rainbow} showed PER to be one of the most critical improvements over DQN. 
\citet{Horgan2018ApeX} reuse the idea of prioritization according to the TD errors in a distributed framework of agents working in parallel.
As a follow-up on PER, \citet{fujimoto2020equivalence} propose new sampling schemes based on the TD errors and adjusted by the loss, whereas \citet{gruslys2018reactor} consider a prioritization on sequences of transitions.

Other methods to select a mini-batch have been proposed. 
To emphasize recent experience and draw a mini-batch of size $B$, \citet{zhang2017deeper} sample uniformly $B-1$ transitions from the replay buffer and add the last experience tuple collected along the trajectory. 
Similarly, \citet{wang2019boosting} have observed a faster convergence of the soft actor-critic algorithm \citep[SAC]{Haarnoja2018SAC} by sampling more frequently the recent experiences collected, which can be seen as a smooth version of the sampling proposed by \citet{zhang2017deeper}. 
\citet{li2019state} try to fill each mini-batch with samples taken from every region of the state-action space $S \times A$, thus emulating a uniform distribution across $S\times A$ within the replay buffer. Conversely to our contribution, where we work on the replay buffer distribution as it is, DisCor \citet{kumar2020discor} selects less often errorful target values, leading to better models.
Despite titles mentioning experience replay, many papers are quite weakly related to the idea of non-uniform sampling of mini-batches. 
For instance, Hindsight Experience Replay \citep{HER2017} is an intrinsic motivation method.
Similarly, \citet{fedus2020revisiting} study the relations between the replay buffer size and the frequency of gradient steps, as well as the crucial importance of n-steps returns, but without considering the question of how to sample mini-batches.

In Supervised Learning, the links between non-uniform sampling of training sets, variance of the stochastic gradient estimate and speed of convergence have notably been studied by \citet{Needell2014SGD,ZhaoZhang2015,wang2017accelerating}. 
As shown therein, the smaller the variance of the gradient estimate, the better the convergence speed. 
Importance sampling \citep{rubinstein2016simulation} can be used to reduce variance while keeping these estimates unbiased. 
Hence, non-uniform sampling schemes can be designed to reduce the variance of the estimate and accelerate the convergence. 
For an SGD update in its simplest form, the ideal sampling scheme $p^*$ is proportional to the per-sample gradient norm \citep{Needell2014SGD}.

\looseness=-1
Computing the optimal sampling distribution requires computing all per-sample gradients. For large training sets (such as RL replay buffers), this task is prohibitively costly. \citet{Alain2016Variance} deploy heavy computational resources to compute the optimal distribution, using clusters of GPUs. Another possibility is to shift from $p^*$, as developed by \citet{Loshchilov2016}, where the sampling is proportional to a loss ranking. Ensuring convergence in convex cases, Stochastic Variance Reduced Gradient \citep{johnson2013accelerating} is a state-of-the art algorithm using importance sampling. Recently, \citet{Idiap2018NASACE} proposed an upper-bound on the per-sample gradient norm which is fast to compute and can be used as a surrogate of $p^*$.

\section{Gradient Variance Minimization}
\label{sec:isavi}

\subsection{Importance Sampling Distributions and Approximations in PER}
 
SGD aims at minimizing the empirical loss, given the current replay buffer, as a proxy for the true loss. 
Hence, at each training step, plain SGD samples uniformly a mini-batch of $B$ transitions from the replay buffer, in order to approximate the gradient of the empirical loss $\frac{1}{N} \sum_{i=1}^N \nabla_\theta \ell\left(Q_\theta\left(x_i\right),y_i\right)$. 
Let $u$ be the uniform discrete distribution over the items in the replay buffer. The gradient can then be written as an expectation over these items: 
\begin{align*}
 \frac{1}{N} \sum_{i=1}^N \nabla_\theta \ell(Q_\theta(x_i), y_i) &= \sum_{i=1}^N u_i \nabla_\theta \ell(Q_\theta(x_i), y_i) \\
 &= \mathbb{E}_{i \sim u} [ \nabla_\theta \ell(Q_\theta(x_i), y_i)]. 
\end{align*}
This expectation can be estimated by an (unbiased) empirical mean over a mini-batch of $B$ samples: 
$$\mathbb{E}_{i \sim u} [\nabla_\theta \ell(Q_\theta(x_i), y_i)] \approx \frac{1}{B} \sum_{i=1}^B \nabla_\theta \ell(Q_\theta(x_i), y_i) \ , \ i\sim u.$$
Let $p$ be any probability distribution over the items of the replay buffer such that $p_i\neq 0, \forall i$. 
Importance sampling 
gives
alternate unbiased estimates of the empirical loss' gradient: 
\begin{align*}
    \mathbb{E}_{i \sim u} [ \nabla_\theta \ell(Q_\theta(x_i), y_i)] &= \mathbb{E}_{i \sim p} \left[ \nabla_\theta \ell(Q_\theta(x_i), y_i)\frac{u_i}{p_i}\right] \\
    &= \frac{1}{N} \mathbb{E}_{i \sim p} \left[ \nabla_\theta \ell(Q_\theta(x_i), y_i)\frac{1}{p_i}\right].
\end{align*}
This %
can also be estimated by an empirical mean: 
$$\mathbb{E}_{i \sim p} \left[\nabla_\theta \ell(Q_\theta(x_i), y_i)\frac{1}{p_i}\right] \approx \frac{1}{B} \sum_{i=1}^B \nabla_\theta \ell(Q_\theta(x_i), y_i) \frac{1}{p_i},$$
$i\sim p.$ The update equation for $\theta$ becomes: 
\begin{align}
    \theta_{t+1} = \theta_t - \eta \frac{1}{B} \sum_{i=1}^B \frac{1}{N p_i}\nabla_{\theta} \ell(Q_\theta(x_i), y_i). & &
    \label{eq:SGD_IS}
\end{align}

\looseness=-1
We define $G_i = w_i \nabla_{\theta} \ell(Q_\theta(x_i), y_i)$ with $w_i = 1/(Np_i)$ for any sampling scheme $p$. 
The gradient of the empirical loss is thus precisely $\mathbb{E}_{i \sim p}[G_{i}] = \sum_{i=1}^N p_i G_i$.
Following the notations of \citet{wang2017accelerating} or \citet{Idiap2018NASACE}, let us define the convergence speed $S$ of SGD under a sampling scheme $p$ as
\begin{equation*}
    S(p) = -\mathbb{E}_{i \sim p} \left[ \|\theta_{t+1} - \theta^*\|_2^2 - \|\theta_t - \theta^*\|_2^2 \right].
\end{equation*}
\citet{wang2017accelerating} show that
\begin{equation*}
    S(p) = 2 \eta (\theta_t - \theta^*)^T \mathbb{E}_{i \sim p}[G_i] - \eta^2 \mathbb{E}_{i \sim p} [G_{i}^T G_{i}].
\end{equation*}
In this work, we call \emph{variance of the gradient estimate} the term $\mathbb{E}_{i \sim p} [G_{i}^T G_{i}]$, which is linked to the covariance matrix $\mathbb{V}\textnormal{ar}_{i \sim p} [G_{i}]$ by $\mathbb{E}_{i \sim p} [G_{i}^T G_{i}] = \textnormal{Tr}( \mathbb{V}\textnormal{ar}_{i \sim p} [G_{i}]) + \mathbb{E}_{i \sim p} [G_{i}]^T \mathbb{E}_{i \sim p} [G_{i}]$. 
It is thus possible to improve the theoretical convergence speed by sampling from the distribution that minimizes $\mathbb{E}_{i \sim p} [G_{i}^T G_{i}]$. 
Indeed, the term $\EE_{i \sim p}[G_i]$ is a constant with respect to $p$ since $\EE_{i \sim p}[G_i] = \sum_{i=1}^N p_i G_i = \sum_{i=1}^N p_i \frac{1}{N p_i} \nabla_{\theta} \ell(Q_\theta(x_i), y_i) = \EE_{i\sim u}[\nabla_\theta \ell(Q_\theta(x_i), y_i)]$, with $u$ the uniform distribution such that $u_i = 1/N$.
The optimal distribution is
\begin{equation*}
    p^*_i \propto \|\nabla_\theta \ell(Q_\theta(x_i), y_i)\|_2,
\end{equation*}
the per-sample gradient norm. 
This derivation is recalled in Appendix \ref{app:optimal}.

The optimal sampling scheme requires computing $\nabla_\theta \ell(Q_\theta(x_i),y_i)$ for all items in the replay buffer, which is too costly to be used in practice. 
Indeed, computing per-sample gradients requires a forward and a backward pass on the whole replay buffer before performing each gradient step. 
Departing from this observation, we first give perspectives to what is done in PER, and then explore two new sampling strategies.

At a given time step, PER computes TD errors as priorities only for the samples selected in the mini-batch, and keeps priorities unchanged for all other samples in the replay buffer. 
We first investigate whether the per-sample gradient norm can be safely approximated by the TD error.
For the sake of notation simplicity, let $q_i$ denote the output $Q_\theta(x_i)$ of the Q-function network.
Then, applying the chain rule to the loss' gradient indicates that the gradient norm is $\|\nabla_\theta \ell(q_i, y_i)\|_2 = \left\| \partial \ell(q_i,y_i) / \partial q_i \cdot \partial q_i / \partial \theta \right\|_2$. 
If $\ell$ corresponds to the $L2$-norm, then $\partial \ell(q_i,y_i) / \partial q_i$ is the TD error. 
Consequently, the per-sample gradient norm of the loss is the product of the TD error and of the norm of the network output's gradient. 
 
Therefore, the TD error is a good approximation of the optimal sampling distribution if $\ell$ is the $L2$-norm and if the norm of $\partial q_i / \partial \theta = \nabla_\theta Q_\theta (x_i)$ is approximately constant across samples $x_i$. 
If one uses the Huber loss instead of the $L2$-norm, then the TD error $\delta_i$ is replaced by $\min(|\delta_i|, 1)$ (which is consistent with the common practice of PER which uses the Huber loss and clips the TD errors). 
However, if the assumption of $\partial q_i / \partial \theta = \nabla_\theta Q_\theta (x_i)$ approximately constant across samples does not hold, then the variance of a sampling according to the TD errors can be higher than that of a uniform sampling.
We provide in Appendix \ref{app:discussion} a simple counter-example demonstrating that the variance induced by the TD error sampling scheme is uncontrolled and potentially even higher than that of a uniform sampling.
 
 \looseness=-1
Besides being approximated by the TD errors, the per-sample gradient norms in PER are also outdated.
Only the samples in the mini-batch receive a priority update at each time step (which already does not correspond to the loss' gradient norm).
All other samples retain priorities %
related to even more ancient Q-functions and are even more outdated.
As such, a priority can be arbitrarily old.
Hence, the variance induced by the sampling scheme used in PER is unknown.

\looseness=-1
As we have just seen, PER uses two approximations. 
We introduce two new sampling schemes, intending to remove these approximations.
We also wish to study which approximation is the most penalizing.
First, we can work on exact but outdated gradients $\nabla_\theta \ell(Q_\theta(x_i), y_i)$, i.e. gradients computed for some past Q-function parameter $\theta_{t'<t}$ and not the current $\theta_t$.
Such gradients are calculated during a backward pass each time the considered sample is selected in a mini-batch. 
We call this strategy GER, for Gradient Experience Replay.
Second, we propose to work with up-to-date priorities.
We call this strategy LaBER, for Large Batch Experience Replay.
Indeed, LaBER 1) pre-samples uniformly a large batch from the replay buffer; 2) computes some importance sampling probability on this large batch; and 3) down-samples the large batch to a mini-batch according to the approximation of the optimal sampling distribution.
LaBER considers the large batch is diverse enough to be representative of the whole replay buffer.
LaBER can either compute the exact gradient norms on the large batch and pay the corresponding computational cost, or exploit some surrogate model to derive approximate sampling probabilities for a reduced cost.
The possible strategies are summarized in Table \ref{tab:possib}.

\begin{table}
  \centering
  \caption{Summary of the main algorithms.}
  \begin{tabular}{l|ll}%
    Priority     & Exact     & Approximate \\
    \hline
    Outdated & GER (this work) & PER \\% \citep{Schaul2015prioritized}     \\
    Up-to-date     & LaBER (this work) & LaBER (this work)      %
  \end{tabular}
  \label{tab:possib}
\end{table}

\subsection{Gradient ER and Large Batch ER}

GER (Algorithm \ref{alg:algorithm1}) shares many similarities with PER: the only differences are the priority value and the hyper-parameters involved in the calculation. 
GER stores per-sample gradient norms whereas PER works with TD errors. 
Both GER and PER work on outdated estimates of the optimal sampling distribution. 
Once an SGD update is performed, $\theta$ changes. 
In turn, the true per-sample gradient norms change, as well as the TD errors, while only the priorities of the samples in the last mini-batch are updated using the latest gradients (which already differ from the true gradients). 
Hence, the variance of the gradient estimates built with PER or GER is uncontrolled, and one cannot guarantee that the convergence speed will be better than with uniform sampling. 
It seems to happen in practice that PER can actually degrade the convergence speed of SGD, as noted for instance by \citet{obando2020revisiting}.
We also emphasize that TD errors are approximations of the per-sample gradient norms. 
Therefore, GER is an attempt at bringing PER closer to the ideal sampling scheme: PER works with outdated and approximated quantities, whereas GER uses outdated but exact quantities. 
Note also that GER incurs the exact same computational cost as PER, since the per-sample gradients on the mini-batches are necessarily computed in the backward pass.

In order to regain control on the gradient's variance, one may wish to approach the up-to-date optimal sampling distribution. Although computing gradient norms on the full replay buffer is prohibitively costly, one can still sample a large batch, compute the gradient norms on this batch and then down-sample it according to the variance reduction distribution. This naturally raises the question of whether this approach brings any benefit compared to plain SGD with larger mini-batches. Section \ref{sec:empirical} and Appendix \ref{app:large} will demonstrate the superiority of importance sampling even in this case. LaBER (Large Batch Experience Replay) is the procedure that samples a large batch of size $mB$ ($m\geq 1$) from the replay buffer, computes the gradients norms, then down-samples it to a mini-batch of size $B$ to perform the SGD update of Equation \eqref{eq:SGD_IS}.
Conversely to PER or GER, whose computing speed can be increased with the use of sum-trees, LaBER does not need specific data-structures. Indeed, LaBER does not sample with a non-uniform distribution the whole replay buffer, but only the large batch, hence tempering the interest of data-structures efficient at large scale.

Computing the optimal variance reduction distribution on the large batch requires running a backward pass on $mB$ items, which is often considered more costly in practice than the forward pass (see Appendix \ref{app:time} for a theoretical discussion and experimental validation of this statement).
As an alternative, we study how one can derive a surrogate model of the gradient norms that can be computed solely via the forward pass.
We follow the notations of \citet{Idiap2018NASACE} and let $z_i$ denote the last layer's input in the $Q_\theta$ network. 
With $\sigma$ the layer's activation function, the network output is $q_i = \sigma\left(z_i\right)$. 
Then, applying the chain rule yields $\|\nabla_\theta \ell(q_i, y_i)\|_2 = \left\| \partial \ell(q_i,y_i) / \partial q_i \cdot  \partial q_i / \partial z_i \cdot \partial z_i / \partial \theta \right\|_2$.
Repeated application of the chain rule leads to the back-propagation algorithm. 
\citet{Idiap2018NASACE} remark that $\left\| \partial \ell(q_i,y_i) / \partial q_i \cdot \partial q_i / \partial z_i  \right\|_2 \left\| \partial z_i / \partial \theta \right\|_2$ is an upper-bound on the per-sample gradient norm of the loss. 
They argue that activation normalization techniques make $\left\| \partial z_i / \partial \theta \right\|_2$ almost constant across samples $x_i$. 
Let $K$ be the largest value of this term.
Thus, the gradient's norm variations across samples should be captured by the other term in the product. 
Write $\Sigma'\left(z_i\right) = \partial q_i / \partial z_i$ the diagonal matrix of activation derivatives $\sigma'(z_i)$ for the last layer, and $\partial \ell(q_i,y_i) / \partial q_i = \nabla_{q_i} \ell(q_i,y_i)$. 
Then $\hat{G}_i = K \left\| \Sigma'\left(z_i\right) \nabla_{q_i} \ell(q_i,y_i) \right\|_2$ 
is an upper bound on 
$\|\nabla_\theta \ell(q_i, y_i)\|_2$. 
When this upper-bound is tight\footnote{The counter-example mentioned in the previous section illustrates why it is not always the case.}, i.e. when $\hat{G}_i$ is an accurate surrogate model of $\|\nabla_\theta \ell(q_i, y_i)\|_2$, it provides a generic way of computing the priority of $x_i$ in a single forward pass, without requiring 
the costly backward pass computation.

Let $\hat{p}$ be the discrete probability distribution over the replay buffer defined by $\hat{p}_i \propto \hat{G}_i$ and $\hat{w}$ the corresponding importance sampling weights. 
Since $\|\nabla_\theta \ell(q_i, y_i)\|_2 \leq \hat{G}_i$, one has a lower bound on the convergence speed of SGD when sampling according to $\hat{p}$:
\begin{equation*}
    S(\hat{p}) \geq 2 \eta (\theta_t - \theta^*)^T \mathbb{E}_{i \sim \hat{p}}[G_i] - \eta^2 \mathbb{E}_{i \sim \hat{p}} [ \hat{w}_i^2 \hat{G}_{i}^2].    
\end{equation*}
The tighter the bound on $\|\nabla_\theta \ell(q_i, y_i)\|_2$, the larger the convergence speed.

\looseness=-1
LaBER with surrogate gradient norms is summarized in Alg. \ref{alg:algorithm2}. %
Conversely to GER, LaBER works with up-to-date, exact or approximate per-sample gradient norms (Table \ref{tab:possib}). 

We remark that a common choice for Q-networks is to use an identity function as the last layer's activation. In this case, the surrogate $\hat{G}_i$ for the per-sample gradient norm boils down to the absolute TD error $|\delta_i|$ when the loss is the $L2$-norm (and to $\min(|\delta_i|, 1)$ for the Huber loss).

Interestingly, GER and LaBER can be combined, since it is possible to sample a large batch according to the distribution proposed by GER, compute the surrogate gradient norms on the large batch, and down-sample according to the computed up-to-date priorities. 
We call this agent GER-LaBER, and since this can also be done with PER, we introduce another agent called PER-LaBER, where the large batch is sampled according to the outdated TD errors stored by PER. Appendix \ref{app:visual} summarizes all the proposed algorithms with a visual representation.

\begin{algorithm}%
\caption{PER and GER}
\label{alg:algorithm1}
\begin{algorithmic}
\REQUIRE Replay buffer, priority list, mini-batch size.
\STATE Priorities := TD errors (PER) or $\|\nabla_\theta \ell(q_i, y_i)\|_2$ (GER).
\LOOP
\STATE Sample a prioritized mini-batch. 
\STATE Compute per-sample gradients. 
\STATE Update priorities. 
\STATE Perform SGD update.
\ENDLOOP
\end{algorithmic}
\end{algorithm}

\begin{algorithm}%
\caption{LaBER with surrogate priorities}
\label{alg:algorithm2}
\begin{algorithmic}
\REQUIRE replay buffer, mini-batch size, large batch size.
\LOOP
\STATE Sample uniformly a large batch.
\STATE Compute surrogate priorities (e.g. TD errors).
\STATE Down-sample according to surrogate priorities.
\STATE Compute per-sample gradients on mini-batch.
\STATE Perform SGD update on mini-batch.
\ENDLOOP
\end{algorithmic}
\end{algorithm}

\subsection{Scaling the Descent Directions}

Given a sampling distribution $p$, one can use Equation \eqref{eq:SGD_IS} to perform SGD updates.
By using $p_i = \hat{G}_i / \sum_{j=1}^N \hat{G}_j$ in Equation \eqref{eq:SGD_IS}, the gradient estimate becomes:
$$\frac{1}{B}\frac{\sum_{j=1}^N \hat{G}_j}{N} \sum_{i=1}^B \frac{1}{\hat{G}_i}\nabla_{\theta} \ell(Q_\theta(x_i), y_i). $$
 
A possible issue lies in the fact that the average gradient norm $\sum_{j=1}^N \hat{G}_j / N$ varies from one SGD update to the other and thus computing it would require computing all the per-sample gradient surrogates. 
Had it been fixed, it could have been ignored and integrated in the learning rate; but since it varies across updates, it affects the SGD step-size and deserves some attention. 
To the best of our knowledge, this issue has not been tackled in the importance sampling attempts at SGD in Supervised Learning. 
We propose three approximation procedures to account for this aspect which we call LaBER-mean, LaBER-lazy, and LaBER-max.

LaBER-mean approximates the average gradient norm over the replay buffer by the average over the large batch. The descent direction is thus: 
$$\frac{1}{B}\frac{\sum_{j=1}^{mB} \hat{G}_j}{mB} \sum_{i=1}^B \frac{1}{\hat{G}_i}\nabla_{\theta} \ell(Q_\theta(x_i), y_i).$$

LaBER-lazy avoids the problem altogether and neglects the variations of the average gradient norm across SGD steps. The constant multiplicative term $\sum_{j=1}^N \hat{G}_j / N$ is absorbed by the learning rate and the descent direction becomes:
$$\frac{1}{B} \sum_{i=1}^B \frac{1}{\hat{G}_i}\nabla_{\theta} \ell(Q_\theta(x_i), y_i).$$

\citet{Schaul2015prioritized} dedicate a section to the study of the weights in the SGD update for PER, and propose to divide the $1 / (Np_i)$ weights which should normally be used, by the maximum weight encountered in the selected mini-batch.
The normalization factor is thus $(\max_{j \in [1,B]}1/\hat{G}_j)^{-1} = \min_{j \in [1,B]} \hat{G}_j$.
Following this common practice for PER, we propose a LaBER-max agent using the descent direction:
\begin{equation}
    \frac{1}{B} \left( \min_{j \in [1, B]} \hat{G}_j \right)
    \sum_{i=1}^B \frac{1}{\hat{G}_i} \nabla_{\theta} \ell(Q_\theta(x_i), y_i).
    \label{eq:LaBER_MAX}
\end{equation}

Note that only the normalization of LaBER-mean is theoretically grounded since it converges to the ideal normalization when $mB$ tends to the full replay buffer size. The two other normalization schemes do not enjoy this property.

\subsection{Extension to Distributional RL and Actor-Critics}%

For the sake of clarity, we exposed the reasoning in this section on the DQN case, but the same considerations apply to the application of any Bellman operator. 
In particular, it holds for the evaluation equation and for the distributional counterparts of the optimality and evaluation equations. 
In the distributional context, for the C51 agent \citep{bellemare2017distributional}, the surrogate sampling scheme obtained via $\hat{G}$ is proportional to the $L2$-norm of the vector containing the 51 per-atom TD errors (proof in Appendix~\ref{app:C51}). 
This contrasts with the combination of C51 and PER defined in Rainbow \citep{Hessel2018Rainbow}, where the priorities are arbitrarily defined as the per-sample losses. 
We note that this might be a poor approximation of the per-sample gradient norm.
The surrogate distribution can also be derived for other distributional agents, such as QR-DQN \citep{dabney2018distributional} or IQN \citep{dabney2018implicit}. For actor-critic algorithms with two critics~\citep{fujimoto2018addressing}, the surrogate distribution is straightforward and should be computed separately on each critic. 
Details are provided in Appendix \ref{app:dopamine}.

\section{Empirical evaluation}
\label{sec:empirical}

\begin{figure}[tbh]
\includegraphics[width=\linewidth]{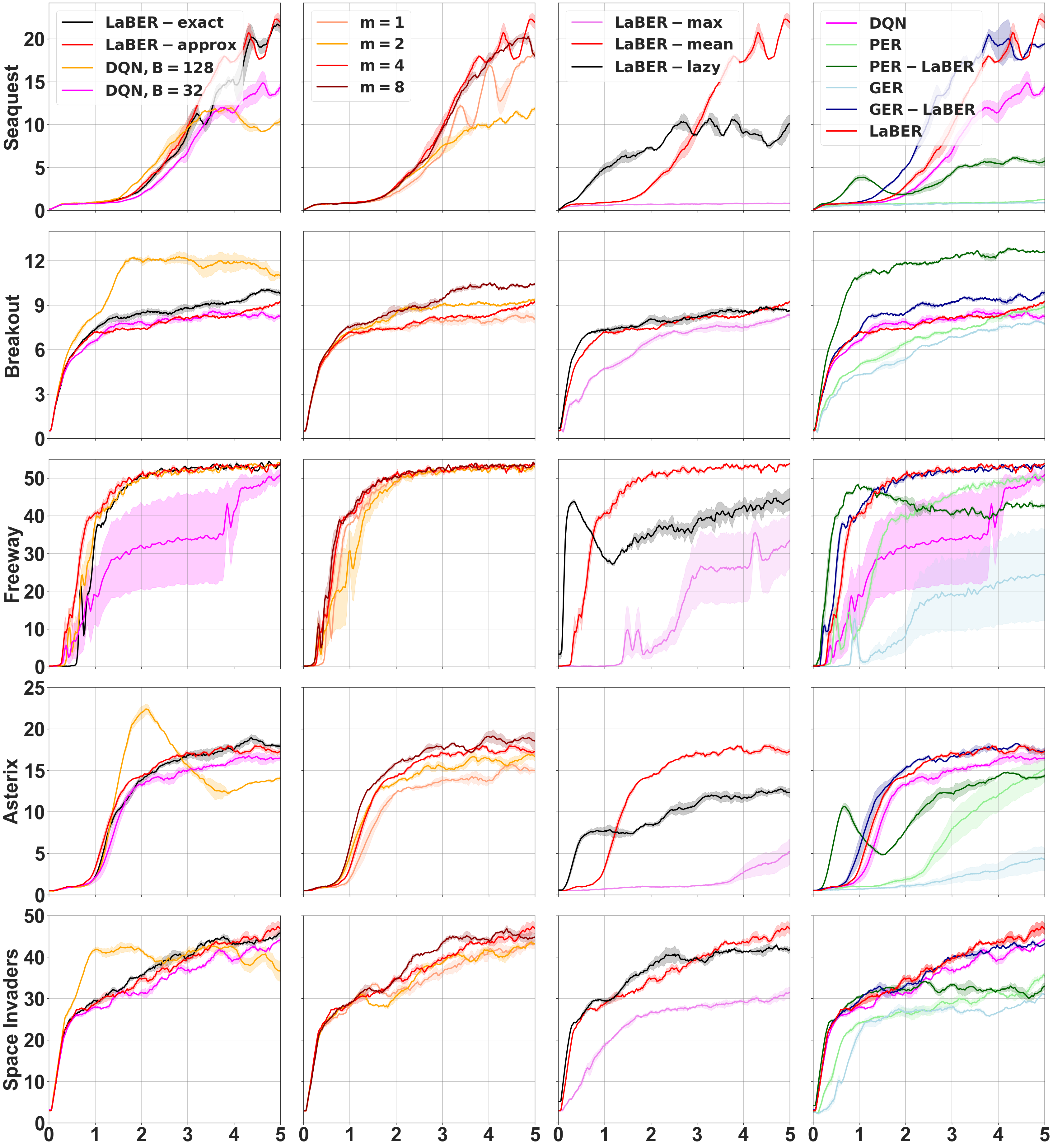}
\caption{The first column compares the LaBER  agent (LaBER-mean with $m=4$) with exact and approximate gradient norms, and DQN with conventional and large mini-batch size ($B$ and $4B$) on MinAtar games. The second column compares different values of $m$ for the LaBER-mean agent. The third column compares the three flavors of LaBER with $m=4$. Finally, the last column compares all the studied algorithms. The x-axis is the number of interaction steps in millions. The y-axis is the average sum of rewards gathered along each episode.}
\label{fig:minatar}
\end{figure}

\begin{figure}[tbh]
\includegraphics[width=0.95\linewidth]{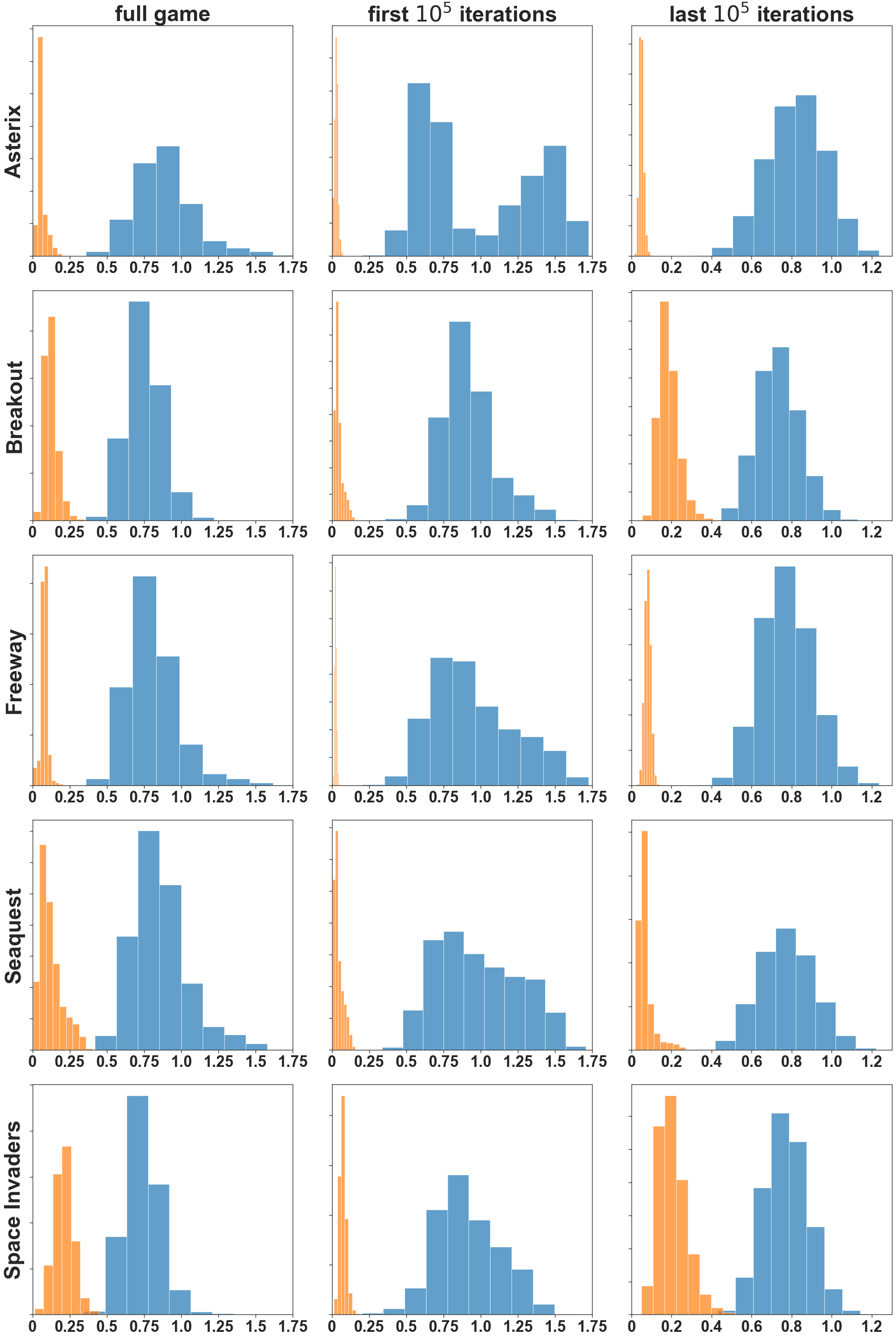}
\caption{The first column reports the histogram of the TVs encountered during training on MinAtar games. The second column uses the TVs computed during the first $10^5$ iterations, and the third column, the last $10^5$ iterations. The TVs between the surrogate and the optimal distributions are in orange, the TVs between the uniform and the optimal distributions are in blue. }
\label{fig:minatarTVmain}
\end{figure}

This section provides a practical analysis of the sampling schemes proposed earlier. In particular, we isolate various aspects of the sampling schemes in order to assess what is really decisive in practice. Notably, the main finding is that working with up-to-date priorities is the key feature for better performance. Hence, from a practical standpoint, LaBER is the key contribution of this paper.

To ease reproducibility, we make our code available at \url{https://github.com/sureli/laber} and recall all hyperparameters in Appendix \ref{app:baselines}.
We emphasize that all baseline algorithms have been used with default hyperparameters. 
For each experiment we ran $n$ independent simulations and reported the average and the standard deviation, where $n=3$ for Atari games \citep{bellemare2013arcade}, $n=6$ for MinAtar games \citep{young19minatar} and $n=9$ for Pybullet environments \citep{coumans2019}.
The computing resources used for these experiments are summarized in Appendix \ref{app:compute}.
All presented results can be found with additional details in Appendix \ref{app:gap}, \ref{app:large}, \ref{app:minatar}, and \ref{app:dopamine}.

\textbf{Is $\hat{p}$ far from $p^*$?}
To assess whether the surrogate distribution is indeed close to the optimal one, we first compared LaBER\footnote{Here we used LaBER-mean with $m=4$.} using exact gradient norms and LaBER using surrogate gradient norms. 
This experiment uses MinAtar games \citep{young19minatar}, an RL benchmark simplifying Atari games and aiming at reproducibility while preserving their main difficulties as well as a diversity of situations. 
As can be seen on the first column of Figure \ref{fig:minatar}, the performance of these two agents are very similar. 
Second, we computed the total variation (TV) metric at each time step, between the optimal and the surrogate distributions, over the conventional 5 million training steps. 
These results are reported on Figure \ref{fig:minatarTVmain} and confirm that $\hat{p}$ is a close approximation of $p^*$. 
Appendix \ref{app:gap} provides additional details.
This justifies the use of the surrogate sampling distribution, instead of the computation of all gradient norms within the large batch. 
This saves computing resources without compromising the performance of LaBER.
In the rest of this section, LaBER refers to LaBER with surrogate gradient norms.

\textbf{LaBER outperforms DQN with larger mini-batches.} 
For a fair comparison, DQN agents with mini-batches as large as the large batch of LaBER were compared with LaBER-mean with $m=4$ (Figure \ref{fig:minatar}, first column). 
This experiment has also been done on Atari games in Appendix~\ref{app:large}. 
Overall, LaBER outperforms DQN both in performance and in computing time, confirming that non-uniform sampling is critical to performance and yields better results than uniform sampling with larger mini-batches. Detailed results are provided in Appendix \ref{app:large} and \ref{app:time}.

\textbf{What size for the large batch?} 
Compared to GER and PER, LaBER has only one hyperparameter, namely the size $mB$ of the large batch. 
The second column of Figure \ref{fig:minatar} compares the different values of $m$ for LaBER-mean and a mini-batch size $B=32$, on all MinAtar games. 
With a few exceptions, this intuitive rule holds: the larger the better. 
Indeed, the larger the large batch, the more diverse and representative of the replay buffer it is. 
In all experiments reported below, $m=4$.

\textbf{Descent direction normalization.} The third column of Figure \ref{fig:minatar} reports the comparison between LaBER-mean, LaBER-lazy and LaBER-max, with $m=4$.
LaBER-lazy has a tendency to learn fast at the beginning of the training. 
We conjecture this is due to a larger norm of the gradient steps. 
In turn, this larger norm might be the cause of the observed larger variance in performance.
LaBER-max, which draws inspiration from the scaling used in PER, consistently performs worse than the two other flavors of LaBER, displaying a large variability on Freeway and an inability to learn on Seaquest. 
This raises the question of the relevance of the update equation of PER. 
This comparison justifies the use of LaBER-mean in the experiments discussed below.

\textbf{LaBER increases performance but also reduces performance variance.} 
As explained in Section \ref{sec:isavi}, the sampling used by PER and GER yields gradient estimates with an uncontrolled variance. 
One cannot guarantee a variance reduction in the gradient estimate (and a convergence speed-up) compared to the uniform sampling. 
The experimental results reported on the last column of Figure \ref{fig:minatar} confirm this variability of PER and GER from one game to another.
This result is corroborated by \citet{obando2020revisiting}, who report adding PER to DQN on MinAtar games did not lead to improvements when no specific hyperparameter tuning is performed. 
Appendix \ref{app:PER_GER} features an in-depth discussion about the differences in the priority calculation in PER and GER, especially the arbitrary scaling choices made and the hyperparameters used.
Conversely, LaBER appears to yield policies with a low variance in performance from one experiment to the other.
Overall, LaBER consistently outperforms all other algorithms on MinAtar games, both asymptotically and in terms of variance.

\textbf{Up-to-date priorities are the key.} 
We finally studied the performance of the GER-LaBER and PER-LaBER agents. 
PER-LaBER is often better than PER, and GER-LaBER always better than GER. 
This proves the benefits of using LaBER, whatever the sampling distribution of the replay buffer. 
However, neither PER-LaBER nor GER-LaBER bring a significative improvement over vanilla LaBER, while they come with additional computational burden (both for computing and for storing the priorities).

\begin{figure}[tbh]
\begin{center}
\includegraphics[width=\linewidth]{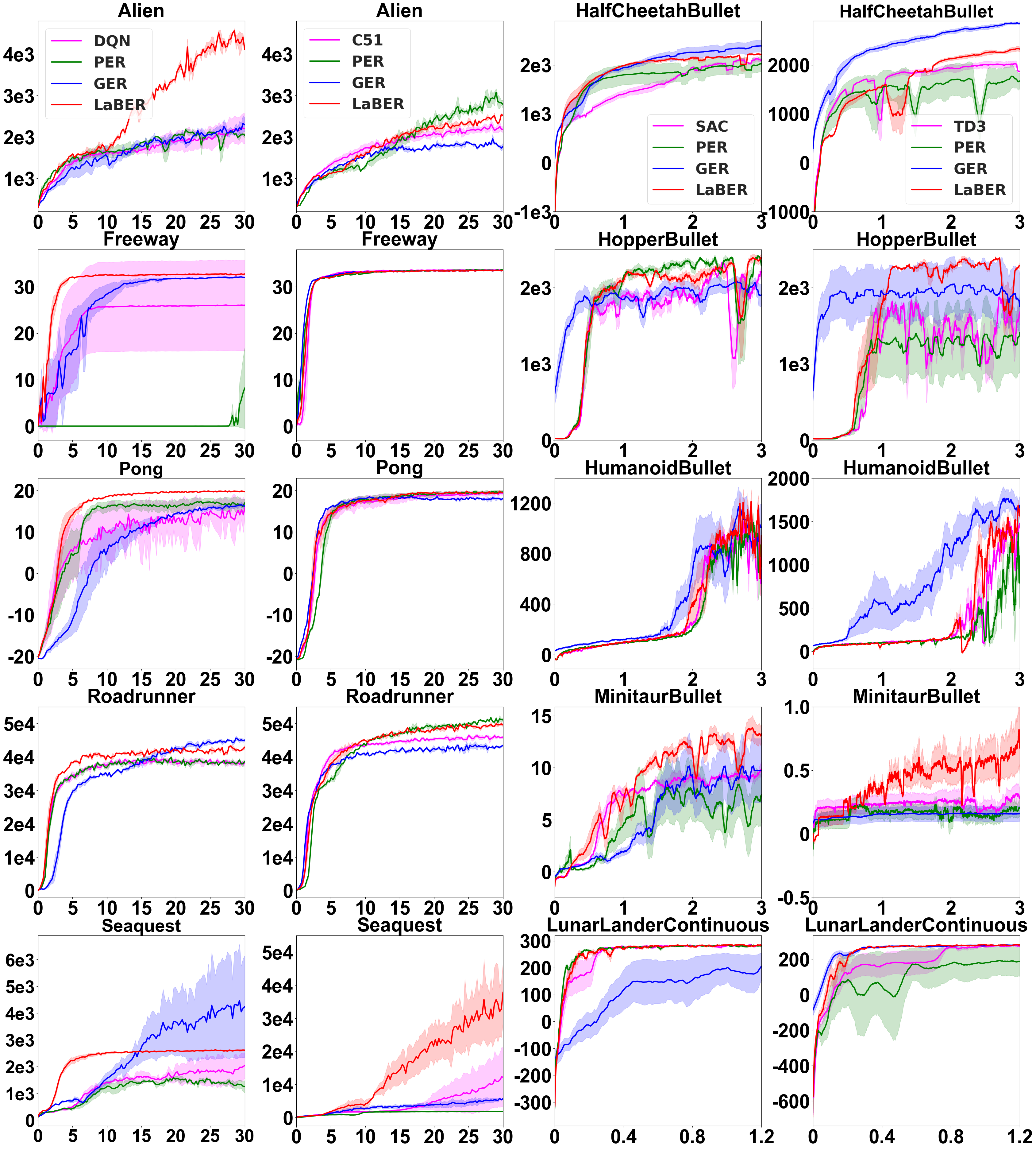}
\caption{\looseness=-1 The $1^\text{st}$ (resp. $2^\text{nd}$, $3^\text{rd}$ and $4^\text{th}$) column compares the studied algorithms with DQN (resp. C51, SAC and TD3). The x-axis is the number of interaction steps in millions. The y-axis is the average Monte Carlo return computed every 250000 steps for Atari games and every 10000 steps for continuous control tasks.}
\label{fig:dopamine}
\end{center}
\end{figure}

\textbf{Distributional RL and actor-critic methods on larger scale benchmarks.} 
Figure \ref{fig:dopamine} reports the results of LaBER, GER and PER when combined with DQN, C51, SAC and TD3 on larger state-space MDPs (namely, 5 games drawn from the Arcade Learning Environment \citep{bellemare2013arcade} and 5 continuous control tasks \citep{coumans2019}).
For Atari games, we used the Dopamine \citep{castro2018dopamine} DQN and C51 implementations as baselines, upon which we implemented prioritization.
For continuous control tasks, our extensions have been implemented over the SAC and TD3 agents provided by Stable-Baselines3 \citep{stable-baselines3}. 
Overall, PER and GER cannot ensure constant improvements without fine hyperparameters tuning, while LaBER consistently remains better than the base agents provided by Dopamine or Stable-Baselines3.

\section{Conclusion}
\label{sec:discussion}

In this work, we have cast the replay buffer sampling problem as an importance sampling one, and shown the link between the theoretical convergence properties of the RL agent and the sampling scheme used. 
Since the optimal sampling distribution is intractable, we made several approximations to compute relevant sampling schemes. 
In particular, we opposed outdated, exact priorities (GER) to up-to-date, exact or approximate ones (LaBER) and demonstrated how they can be combined. 
On the way, we demonstrated that PER can be seen as an importance sampling scheme using outdated and approximated per-sample gradient norms. 
Our theoretical contribution consists in providing sound foundations to algorithms that perform non-uniform sampling in the replay buffer, such as PER.
LaBER is our key algorithmic contribution. It is an easy-to-code algorithm that brings consistent improvement over a significant range of classical RL benchmarks.

\bibliography{biblio}
\bibliographystyle{icml2022}

\newpage
\appendix
\onecolumn

\section{Convergence Speed and Optimal Sampling Distribution}
\label{app:optimal}

From one sampling scheme to another, the variance of the gradient estimate is different, and we are looking for the optimal sampling scheme $p^*$ over items of the replay buffer, the one with smallest variance. 
Indeed, for such a sampling scheme, the convergence speed of SGD is optimum. 
We define the convergence speed $S$ for a sampling scheme $p$ as $S(p) = -\mathbb{E}_{i \sim p} \left[ \|\theta_{t+1} - \theta^*\|_2^2 - \|\theta_t - \theta^*\|_2^2 \right]$.
We recall that a stochastic gradient descent update has the form $\theta_{t+1} = \theta_t - \eta G_i$, where $G_i$ is the gradient estimate built from sampling element $i$ with probability $p$.
The following derivations from \citep{wang2017accelerating} shed light on the relationship between variance and convergence speed:
\begin{align*}
S(p) &= -\mathbb{E}_{i \sim p} \left[ \|\theta_{t+1} - \theta^*\|_2^2 - \|\theta_t - \theta^*\|_2^2 \right] \\
&= -\mathbb{E}_{i \sim p} \left[ \theta_{t+1}^T \theta_{t+1} - 2 \theta_{t+1}^T \theta^* -  \theta_{t}^T \theta_{t} +2 \theta_{t}^T \theta^* \right] \\
&= -\mathbb{E}_{i \sim p} \left[ (\theta_t - \eta G_{i})^T (\theta_t - \eta G_{i}) + 2\eta G_{i}^T \theta^* - \theta_t^T \theta_t \right] \\
&= -\mathbb{E}_{i \sim p} \left[ -2 \eta (\theta_t - \theta^*)^T G_{i} + \eta^2 G_{i}^T G_{i} \right] \\
&= 2 \eta (\theta_t - \theta^*)^T \mathbb{E}_{i \sim p}[G_i] - \eta^2 \mathbb{E}_{i \sim p} [G_{i}^T G_{i}]
\end{align*}
It is possible to gain a speed-up by sampling from the distribution that minimizes $\mathbb{E}_{i \sim p} [G_{i}^T G_{i}]$. This yields the constrained optimization problem:
\begin{gather*}
    \min_{p} \mathbb{E}_{i \sim p}[G_{i}^T G_i] = \min_{p} \sum_{i=1}^N p_i \|G_{i}\|_2^2 \\ \textrm{such that } \sum_{i=1}^N p_i = 1 \textrm{ and } p_i \geq 0
\end{gather*}
Recall that $G_i = w_i \nabla_{\theta} \ell(Q_\theta(x_i), y_i)$ and $w_i = 1/(Np_i)$. Let $g_i = \| \nabla_\theta \ell (Q_\theta(x_i), y_i)\|_2$. The problem boils down to:
\begin{gather*}
     \min_p \frac{1}{N^2} \sum_{i=1}^N \frac{1}{p_i} g_i^2, \\
     \textrm{such that } \sum_{i=1}^N p_i = 1 \textrm{ and }p_i\geq 0.
\end{gather*}

\begin{lemma}[Optimal sampling distribution] 
The optimal sampling distribution $p^*$ verifies $p^*_i \propto \|\nabla_\theta \ell (Q_\theta(x_i), y_i)\|_2$, the per-sample gradient norm.
\end{lemma}

\begin{proof}
We note $\mu \in \mathbb{R}$ the Lagrange multiplier associated to the equality constraint, $\nu \in \mathbb{R}_+^N$ the Lagrange multipliers associated to the inequality constraints. Hence:
\begin{equation*}
    \textnormal{Lag}(p, \mu, \nu) = \sum_{i=1}^N \frac{1}{p_i} g_i^2 + \mu \left(\sum_{i=1}^N p_i-1 \right) -\sum_{i=1}^N \nu_i p_i
\end{equation*}
Setting the derivatives of the Lagrangian with respect to the primal variables yields: 
\begin{equation*}
    \forall \ i \in [ 1, N ], \ -\frac{g_i^2}{p_i^2} + \mu - \nu_i = 0
\end{equation*}
Multiplying the above equation by $p_i$ and using $\forall \ i, \ p_i\nu_i = 0$ (complementary slackness), we have: $p_i = g_i / \sqrt{\mu}$, which yields the result.
\end{proof}

\section{Visual Summary of the Studied Agents for the DQN Case}
\label{app:visual}

\begin{figure}
\begin{center}
\includegraphics[scale=0.35]{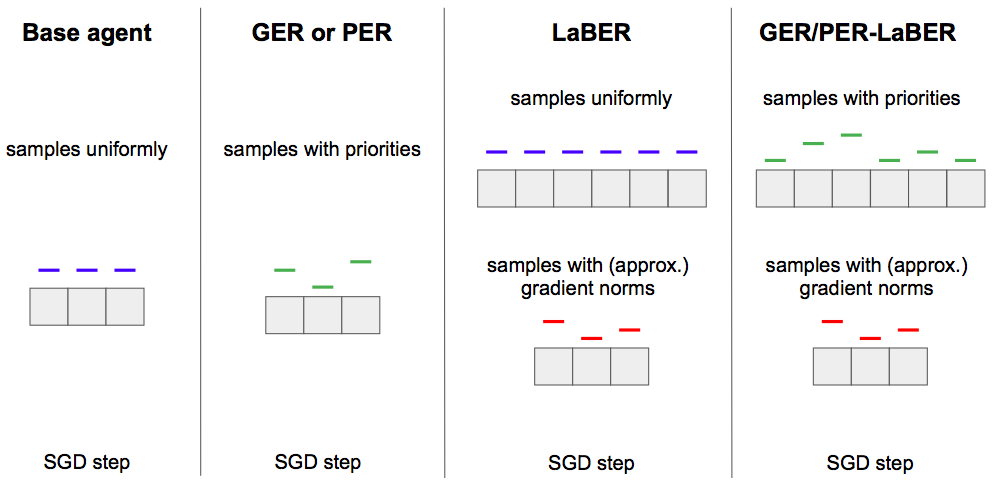}
\caption{Visual representation of the proposed agents for the DQN case}
\label{fig:explanations}
\end{center}
\end{figure}

Figure \ref{fig:explanations} suggests a visual summary of the algorithms studied in this work. 
Grey squares in Figure \ref{fig:explanations} represent experience samples. 
Adjacent squares form a small batch sampled from the replay buffer.
The base algorithm we consider uses uniform sampling.
PER and GER sample according to pre-existing (outdated) priorities on the whole replay buffer. 
LaBER samples uniformly a large batch, computes exact or surrogate gradient norms on this large batch, and finally down-samples the large batch to a mini-batch according to the computed priorities. 
Note that the large batch can be sampled according to the priorities stored by PER or GER, yielding the PER-LaBER or GER-LaBER agents.

\section{Discussion on the Surrogate Distribution}
\label{app:discussion}

In this section, we introduce the counter-example mentioned in Section \ref{sec:isavi}.
The sampling probabilities $p_i \propto \|\Sigma'(z_i)\nabla_{q_i} \ell(q_i, y_i)\|_2$ become $p_i \propto |Q_\theta(x_i) - y_i|$ in the DQN case or in the critic evaluation of an actor-critic method, because the last layer's activation function is the identity function and the $L2$-loss is used. 
We provide an example for $N = 2$ and $B = 1$ where sampling according to the TD errors yields a variance in the gradient estimate that is larger than the variance obtained with the uniform sampling scheme $u$. Let 
\begin{align*}
\left \{
\begin{array}{rcl}
\|\nabla_\theta \ell(Q_\theta(x_1), y_1)\|_2 &=& 10 \\
|Q_\theta(x_1)-y_1| &=& 1 \\
\|\nabla_\theta Q_\theta(x_1)\|_2 &=& 10 
\end{array}
\right. \ \textnormal{and} \
\left \{
\begin{array}{rcl}
\|\nabla_\theta \ell(Q_\theta(x_2), y_2)\|_2 &=& 5 \\
|Q_\theta(x_2)-y_2| &=& 4 \\
\|\nabla_\theta Q_\theta(x_2)\|_2 &=& 5/4. 
\end{array}
\right.
\end{align*}
We find $p^*_1 = 2/3$, $p_2^* = 1/3$, $p_1 = 1/5$ and $p_2 = 4/5$. 
We re-use $G_i = w_i \nabla_{\theta} \ell(Q_\theta(x_i), y_i)$ with $w_i = 1/(Np_i)$ for any sampling scheme $p$, introduced in Section \ref{sec:isavi}. 
To assess the variance of such an estimate, we are interested in the quantity:
$$ \EE_{i \sim p}[G_i^T G_i] = \frac{1}{N^2}\sum_{i=1}^N  \frac{1}{p_i} \|\nabla_\theta \ell(Q_\theta(x_i), y_i)\|_2^2.$$
For the uniform sampling scheme, $u_i = 1/2$. It yields $ \EE_{i \sim u}[G_i^T G_i] = 62.5$. For the ideal sampling scheme $p^*$, it yields $ \EE_{i \sim p^*}[G_i^T G_i] = 56.25$. When sampling according to the $p$, we obtain $ \EE_{i \sim p}[G_i^T G_i] = 132.8125$, which is larger than the variance under the uniform sampling scheme. 

Note that, $p=p^*$ if $\|\nabla_\theta Q_\theta(x_i)\|_2$ is constant. 
As illustrated in this toy example, high variations of $\|\nabla_\theta Q_\theta(x_i)\|_2$ play a crucial role, and can lead to catastrophic increase of the variance. 
In practice, such extreme examples seem to be rare. 
\citet{Idiap2018NASACE}  argue that the per-sample gradient norm variations are mostly caused by the very last layer, reducing the importance of the variations of $\|\nabla_\theta Q_\theta(x_i)\|_2$. 

\section{LaBER for the C51 Agent}
\label{app:C51}

The C51 agent is reputed to perform better than DQN since it does not only learn the Q-function, defined as the expected cumulative sum of rewards, but the whole probability distribution of this sum~\citep{dabney2018distributional}. 
For the $i$th transition tuple $(s_i, a_i, r_i, s_i')$, the neural network outputs a probability distribution under the form of a histogram over 51 atoms, each written $(q^{(k)}_i)_{1 \le k \le 51}$, and learnt by minimizing the Kullback-Leibler divergence with respect to a target histogram probability distribution $(y^{(k)}_i)_{1 \le k \le 51}$. 
The loss for the transition tuple $(s_i, a_i, r_i, s_i')$ can be written:
$$ \ell_i = - \sum_{k=1}^{51} y^{(k)}_i \log q^{(k)}_i,$$ 
and we recall that the last activation of such a network is a softmax function. 

We need to compute the derivative of a softmax loss function to find the surrogate sampling distribution. 
Let $z^{(k)}_i$ be the $k$th element of the input to the softmax layer, the following result holds:
$$\frac{\partial \ell_i}{\partial z^{(k)}_i} = q^{(k)}_i - y^{(k)}_i.$$
Since the gradient is the vector of per-component derivatives, the surrogate of the optimal sampling distribution is: $$p_i \propto \left( \sum_{k=1}^{51} \left(q^{(k)}_i - y^{(k)}_i\right)^2 \right)^{1/2},$$
which is nothing but the $L2$-norm of the vector containing what could be seen as the equivalent of the TD error on each atom.  
Hence, LaBER differs from what is proposed by \citet{Hessel2018Rainbow} for the Rainbow agent, who consider a prioritization on the loss. 

\section{Baseline Agents}
\label{app:baselines}

We denote neural networks structures as follows. ``$\textnormal{Conv}^d_{a,b} \ c$'' is a 2D convolutional layer with $c$ with $c$ feature maps whose kernels have size $a \times b$ and of stride $d$. ``FC $n$'' is a fully connected layer with $n$ neurons.

\subsection{DQN Atari}

The parameters used in our experiments are those reported in the Dopamine reference paper \citep{castro2018dopamine} and recalled in Table \ref{tab:dqn_atari}. We used the provided DQN agent, without modifications. 

\begin{table}
  \caption{DQN parameters for Atari}
  \label{tab:dqn_atari}
  \centering
\begin{tabular} { 
   m{8em} 
   m{29.5em}  }
 \hline
 Parameter & Value \\
 \hline
 Discount factor ($\gamma$)  & 0.99 \\
 Mini-batch size & 32 \\
 Replay buffer size & $10^6$ \\
 Target update period & 8000 \\
 Interaction period & 4 \\
 Random actions rate & 0.01 (with a linear decay of period $2.5 \times 10^5$ steps) \\
 $Q$-network structure & $\textnormal{Conv}^4_{8,8} \ 32 - \textnormal{Conv}^2_{4,4} \ 64 - \textnormal{Conv}^1_{3,3} \ 64 - \textnormal{FC} \ 512 - \textnormal{FC} \ n_A$  \\
 Activations & ReLU (except for the output layer) \\
 Optimizer & RMSProp ($\textit{lr}$: 0.00025, Smoothing constant: 0.95, Centered: True, Epsilon: $10^{-5}$) \\
\hline
\end{tabular}
\end{table}

We follow the procedures of \citet{machado2018revisiting} to train agents in the ALE (Arcade Learning Environment). Notably, we perform one training step (a gradient descent step) every 4 frames encountered in the environment. The state of an agent is the stack of the last 4 frames, sub-sampled to a shape of (84, 84), in shades of grey. We refer to \citet{machado2018revisiting} for details on the preprocessing.

\subsection{DQN MinAtar}

Once again, the parameters are those reported in the MinAtar reference paper \citep{young19minatar} and recalled in Table \ref{tab:dqn_minatar}. We used the provided DQN agent, without modifications.

\begin{table}
  \caption{DQN parameters for MinAtar}
  \label{tab:dqn_minatar}
  \centering
\begin{tabular} { 
   m{8em} 
   m{29.5em}  }
 \hline
 Parameter & Value \\
 \hline
 Discount factor ($\gamma$)  & 0.99 \\
 Mini-batch size & 32 \\
 Replay buffer size & $10^5$ \\
 Target update period & 1000 \\
 Random actions rate & 0.1 (with a linear decay of period $10^5$ steps) \\
 $Q$-network structure & $\textnormal{Conv}^1_{3,3} \ 16 - \textnormal{FC} \ 128 - \textnormal{FC} \ n_A$  \\
 Activations & ReLU (except for the last layer) \\
 Optimizer & RMSProp ($\textit{lr}$: 0.0001, Smoothing constant: 0.95, Centered: True, Epsilon: $10^{-2}$) \\
\hline
\end{tabular}
\end{table}

\subsection{SAC}

For SAC, we used the default parameters provided by Stable-Baselines3 \citep{stable-baselines3} and recalled in Table \ref{tab:sac}. 

\begin{table}
  \caption{SAC parameters}
  \label{tab:sac}
  \centering
\begin{tabular} { 
   m{15em} 
   m{22.5em}  }
 \hline
 Parameter & Value \\
 \hline
 Discount factor ($\gamma$)  & 0.99 \\
 Mini-batch size & 256 \\
 Replay buffer size & $10^6$ \\
 Target smoothing coefficient: & 0.005 \\
 Policy noise & 0.2 \\
 Entropy target & $-\textnormal{dim}(\mathcal{A})$ \\
 Update actor every & 1 \\
 Network structure (critics and actor) & $ \textnormal{FC} \ 256 - \textnormal{FC} \ 256 - \textnormal{FC} \ \textnormal{dim}(\mathcal{A})$  \\
 Activations & ReLU (except for the last layer) \\
 Optimizer & Adam ($\textit{lr}$: 0.001, Epsilon: 0.00001) \\
\hline
\end{tabular}
\end{table}

\subsection{TD3}

For TD3, we used the default parameters provided by Stable-Baselines3 \citep{stable-baselines3} and recalled in Table \ref{tab:td3}.

\begin{table}
  \caption{TD3 parameters}
  \label{tab:td3}
  \centering
\begin{tabular} { 
   m{15em} 
   m{22.5em}  }
 \hline
 Parameter & Value \\
 \hline
 Discount factor ($\gamma$)  & 0.99 \\
 Mini-batch size & 100 \\
 Replay buffer size & $10^6$ \\
 Target smoothing coefficient: & 0.005 \\
 Policy noise & 0.2 \\
 Noise clip & 0.5 \\
 Update actor every & 2 \\
 Network structure (critics and actor) & $ \textnormal{FC} \ 400 - \textnormal{FC} \ 300 - \textnormal{FC} \ \textnormal{dim}(\mathcal{A})$  \\
 Activations & ReLU (except for the last layer) \\
 Optimizer & Adam ($\textit{lr}$: 0.001, Epsilon: 0.00001) \\
\hline
\end{tabular}
\end{table}

\section{PER and GER}
\label{app:PER_GER}

\begin{figure}
\begin{center}
\includegraphics[width=0.8\textwidth]{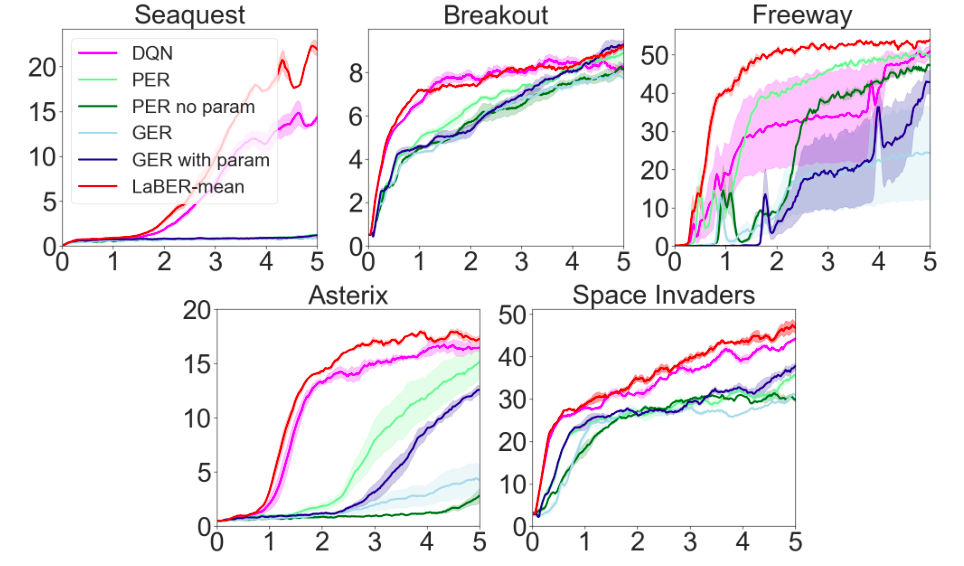}
\caption{PER and GER with different combinations of hyperparameters on MinAtar games. The x-axis is the number of interaction steps in millions. The y-axis is the average sum of rewards gathered along each episode.}
\label{fig:per_ger}
\end{center}
\end{figure}

PER and GER share many similarities, the main difference being the priorities. 
In order to put more emphasis on $(s,a)$-pairs that feature a large approximation error of $T^* Q_n$, PER assigns each transition of the replay buffer a priority based on the TD error $(|\delta_i|+c)^\alpha$, where $\delta_i = Q_\theta(x_i) - y_i$ is the TD error, and with $\alpha$ and $c$ two non-negative hyperparameters.
PER also exponentiates the importance sampling weights in the SGD update equation to the power $\beta$. 
The authors suggest that $c = 10^{-10}$, $\alpha = 0.6$, and $\beta$ growing from $\beta_0$ to 1 are good choices.

In this work, we always consider $\beta = 1$, because it was found to work better on Atari games by the authors of Dopamine \citep{castro2018dopamine}. We used $c = 10^{-10}$ and $\alpha = 0.6$ for PER, but set $\alpha = 1$ and $c = 0$ for GER to work with the exact per-sample gradient norm.

In this section, we study the performance of GER with $\alpha = 0.6$ and $c = 10^{-10}$ as well as PER with $\alpha = 1$ and $c = 0$ on MinAtar games.
In Figure \ref{fig:per_ger},
\begin{itemize}
    \item \textbf{PER} uses $\alpha = 0.6$ and $c = 10^{-10}$;
    \item \textbf{PER no param} uses $\alpha = 1$ and $c = 0$;
    \item \textbf{GER with param} uses $\alpha = 0.6$ and $c = 10^{-10}$;
    \item \textbf{GER} uses $\alpha = 1$ and $c = 0$. 
\end{itemize}

The combination $\alpha = 0.6$ and $c = 10^{-10}$, both for PER and GER, appears to be better than the combination $\alpha = 1$ and $c = 0$. This highlights the sensitivity to hyperparameter tuning when using PER or GER.

\section{Computing Resources}
\label{app:compute}

The experiments presented in this paper exploited internal computing clusters.
The results on Atari games were obtained with single node computations. Each node contained 2 12-core Skylake Intel(R) Xeon(R) Gold 6126 \@ 2.6 GHz CPUs with 96 Go of RAM and 2 NVIDIA(R) Tesla(R) V100 32Go GPUs (only one was used per experiment).
The results on the MinAtar and PyBullet environments used single nodes also. Each of these nodes was composed of 2 12-core Skylake Intel(R) Xeon(R) Gold 6126 \@ 2.6 GHz CPUs with 96 Go of RAM (no GPU hardware).

\section{Assessment of the Gap between $p^*$ and $\hat{p}$}
\label{app:gap}

As explained in the main paper, the experimental performance should be correlated with the quality of the approximate sampling scheme used. 
Comparing the sampling scheme used in PER or GER with the ideal one is prohibitively costly since it requires computing gradient norms on each item in the full replay buffer, at each gradient descent step.
However, for LaBER, the exact per-sample gradient norms can be computed solely on the large batch, and the distribution induced compared with the surrogate model. 
The measure used to compare these distributions is the total variation (TV), defined as $\nu(p,q) = \sum_i |p_i-q_i|$, for two discrete distributions $p$ and $q$.

\begin{figure}
\begin{center}
\includegraphics[width=0.95\textwidth]{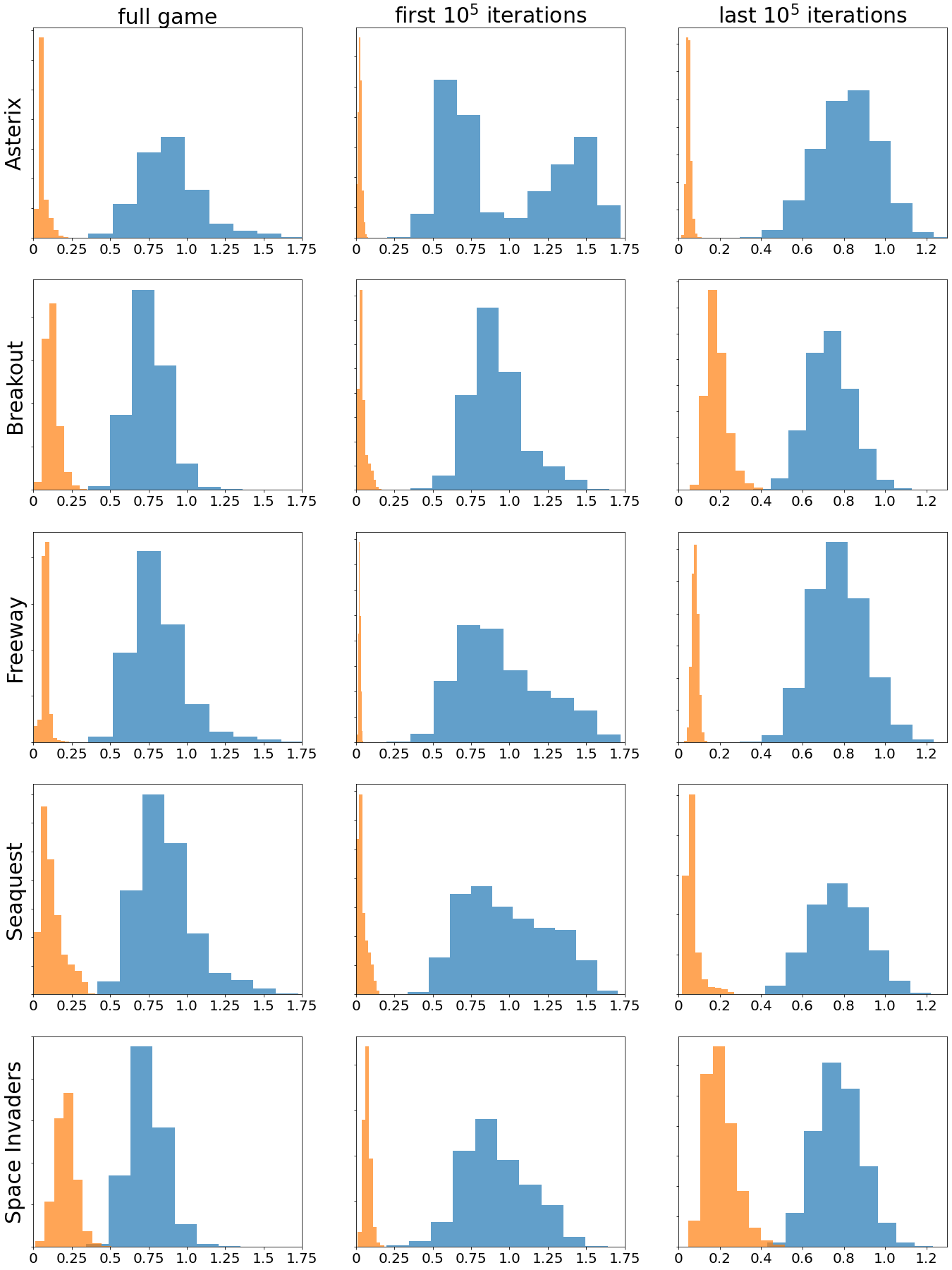}
\caption{The first column reports the histogram of the TVs encountered during training. The second column uses the TVs computed during the first $10^5$ iterations, and the third column, the last $10^5$ iterations. The TVs between the surrogate and the optimal distributions are in orange, the TVs between the uniform and the optimal distributions are in blue. }
\label{fig:minatarTV}
\end{center}
\end{figure}

On the 5 MinAtar games, LaBER-mean with $m=4$ is run during the $5\cdot 10^6$ conventional iterations.
Besides computing the surrogate distribution on the large batch, the distribution induced by the exact per-sample gradient norms on this large batch is also computed at each iteration.
The TV between these two distributions is saved, as well as the TV between the uniform distribution and the one induced by the per-sample gradient norms. 
A histogram of all encountered TVs is then built and reported on Figure \ref{fig:minatarTV}.

On all experiments, the TVs between the uniform and the optimal distributions are statistically significantly higher than the TVs between the surrogate and the optimal distributions, at the beginning, during, and at the end of the experiment. 
The results of these comparisons confirm the accuracy of the surrogate model presented in Section \ref{sec:isavi}.

\section{Comparison between DQN with Larger Mini-Batches and LaBER}
\label{app:large}

\begin{figure}
\begin{center}
\includegraphics[width=0.99\textwidth]{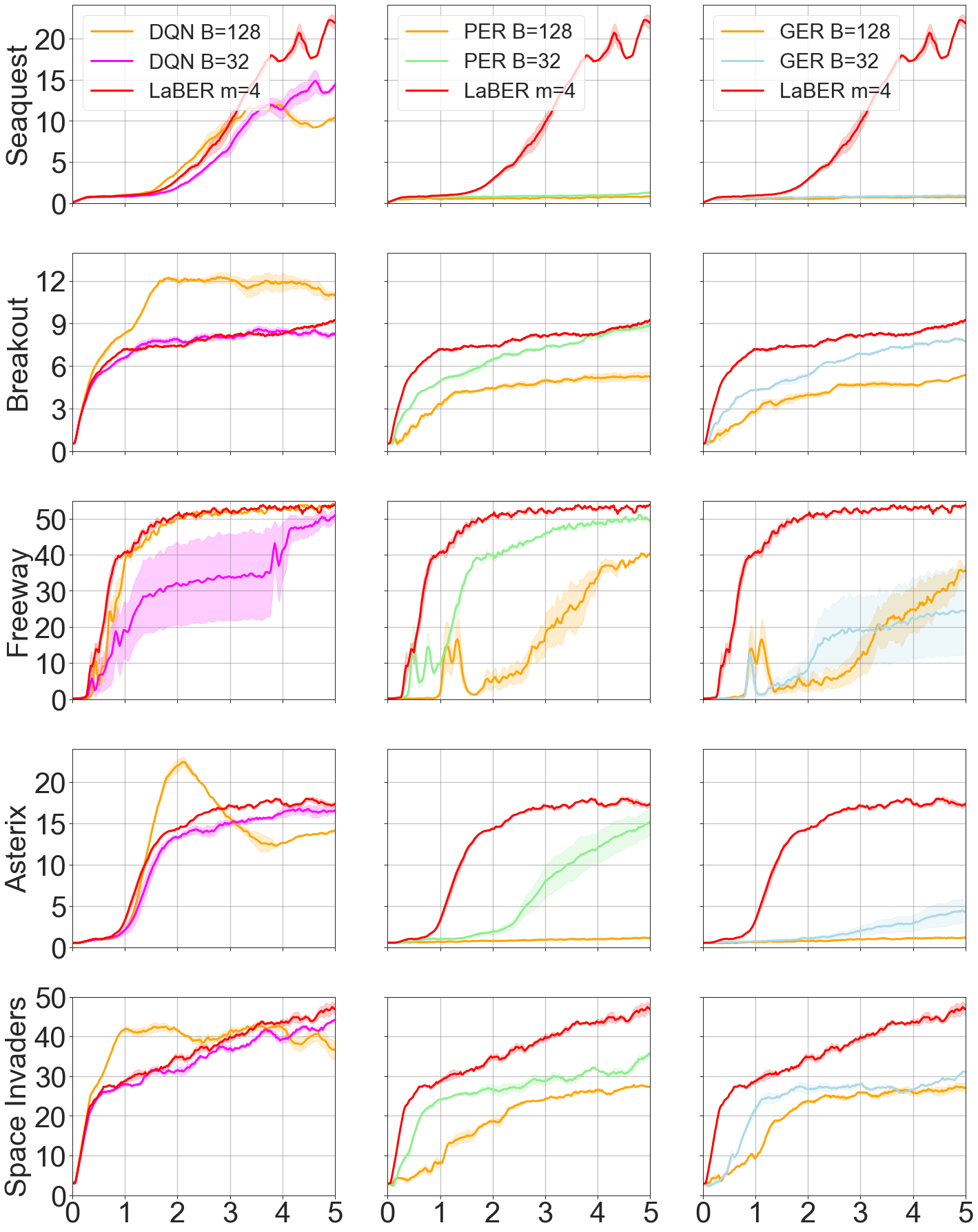}
\caption{On MinAtar games, this figure reports the performance of DQN, PER and GER with large mini-batches ($4B$ instead of $B$) and LaBER with $m=4$. The x-axis is the number of interaction steps in millions. The y-axis is the average sum of rewards gathered along each episode.}
\label{fig:minatarlarge}
\end{center}
\end{figure}

Figure \ref{fig:minatarlarge} reports the performance of DQN, PER, and GER agents with large mini-batches ($4B$ instead of $B$) on MinAtar games, and compares them to LaBER with $m=4$.
This allows confirming that non-uniform samping efficiently reduces variance and outperforms uniform sampling, even when the latter has a larger mini-batch budget.
LaBER-mean, with $m=4$, still outperforms the other agents, 
demonstrating the advantage of a variance reduction importance sampling scheme over uniform sampling with a larger budget, or uncontrolled-variance sampling schemes.

\begin{figure}
\begin{center}
\includegraphics[width=0.99\textwidth]{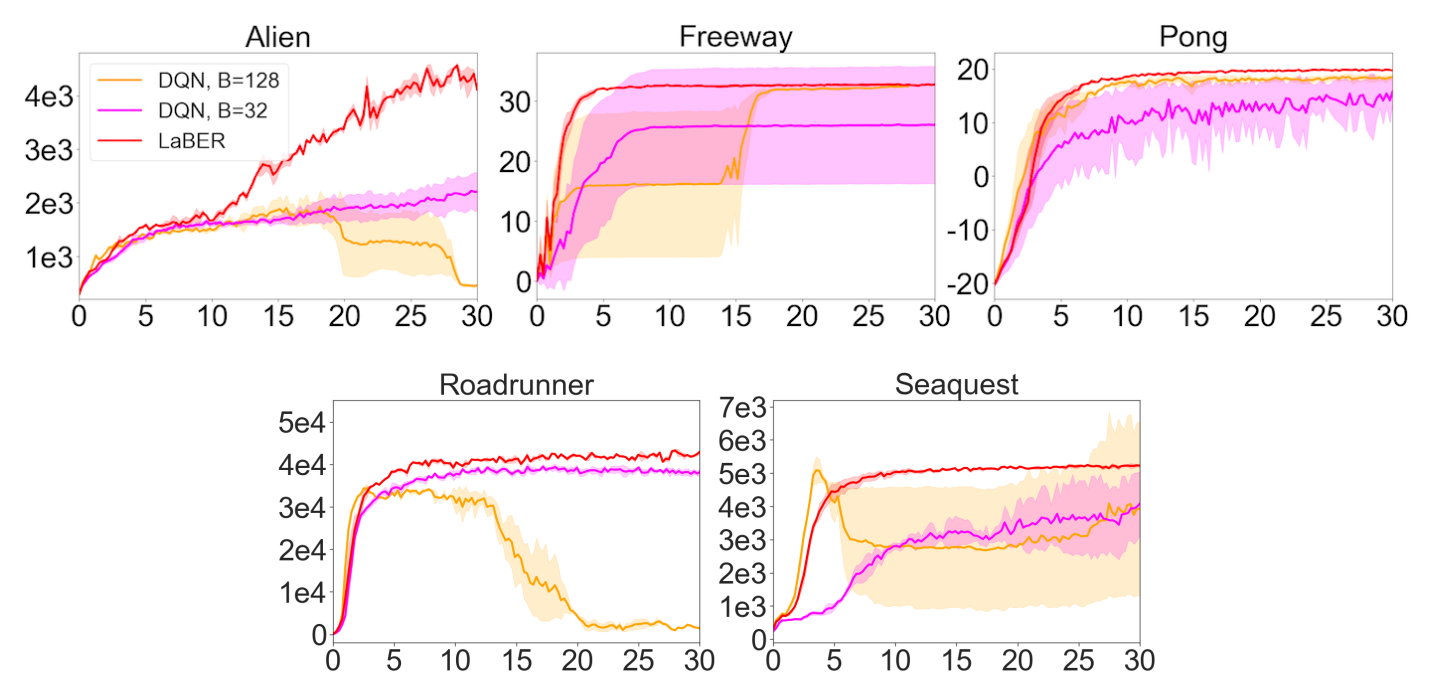}
\caption{DQN with batch size 32 and batch size 128, compared to LaBER-mean with approximate gradient norms with $m=4$, on 5 Atari games. The x-axis is the number of interaction steps in millions. The y-axis is the average Monte Carlo return computed every 250000 steps.}
\label{fig:dopaminelarge}
\end{center}
\end{figure}

Figure \ref{fig:dopaminelarge} completes the first column of Figure \ref{fig:minatar}, on Atari games instead of MinAtar games. LaBER-mean with approximate gradient norms with $m=4$ is better than DQN with a mini-batch as large as the large batch of the LaBER agent.

\section{Cost of Forward and Backward Propagation in Feed-Forward Fully-Connected Neural Networks}
\label{app:time}

This section is a reminder of the computational cost of the forward pass and the backward pass. 
We recall the well-known result that the backward pass is more expensive than the forward pass, in the simple case of a feed forward fully-connected neural network. 
The computational cost of the backward pass justifies the use of LaBER with a surrogate model for the gradient norms, which only adds forward computations. 

Let $\Psi(\cdot, \Theta)$ be a fully connected neural network parameterized by $\Theta$, made of $L$ layers, with $\theta_l$ the weights and biases of layer $l$. 
Let $\sigma_l$ be a Lipschitz continuous activation function. 
Then, the forward pass can be written:
\begin{align}
    x_0 &= x  \notag \\
    z_l &= \theta_l \cdot x_{l-1} \label{eq:fw1}\\
    x_l &= \sigma_l\left(z_l\right) \label{eq:fw2}\\
    \Psi(x, \Theta) &= x_L  \notag
\end{align}

For the sake of simplicity, let $n$ be the number of neurons on each hidden layer, let $B$ be the mini-batch size and let $x_L$ have dimension one. Then Equation \eqref{eq:fw1} induces $O(n^2B)$ operations. Equation \eqref{eq:fw2}, in turn, induces $O(nB)$ operations. Overall, the forward pass induces $O(Ln^2B)$ operations.

Let $\mathcal{L}$ be the loss used to fit the neural network. 
The backward pass, recursively computes the $\nabla_{\theta_l} \mathcal{L} (\Psi(x, \Theta); y)$ gradients for each layer $l$, in order to update the network parameters as:
\begin{align}
    \delta_L &= \sigma_L'(z_L), \notag \\
    \theta_{l} &\leftarrow \theta_l - \eta \left( \delta_l \cdot x_{l-1}^T \right) \nabla_{x_L} \mathcal{L}, \label{eq:bw1} \\
    \delta_{l-1} &= \sigma_{l-1}'(z_{l-1}) \circ \left( \delta_l \cdot \theta_{l-1}  \right), \label{eq:bw2}
\end{align}
where $\circ$ is the Hadamard product. Equation \eqref{eq:bw1} induces $O(n^2B)$ operations. Equation \eqref{eq:bw2} induces $O((n + n^2)B)$ operations. Hence the backward pass induces $O(2Ln^2B)$ operations.

The point of this simple reminder is to recall that the complexity classes of both operations are overall the same, but that the backward pass requires (an order of) twice as many operations as the forward pass. 
This has been repeatedly noticed in the automatic differentiation litterature and reported in neural networks research papers (such as the work of \citet{Idiap2018NASACE} for instance). 
Of course, parallelization and caching in modern computation architectures may greatly amortize this complexity and make it unnoticeable in some cases.  
In the extreme case where parallelization benefits DQN, LaBER with surrogate gradient norms incurs the same wall-clock computation time as DQN.

We verify experimentally this comparison by recording the CPU time taken by the portion of code which differs between DQN and LaBER with surrogate gradient norms. 
The aim of this experiment is to show the benefits of using LaBER from a CPU time point of view. 
Table \ref{tab:time} reports the CPU times of the forward and the backward passes for plain DQN with different values of batch size and LaBER with different values of $m$. 
These results have been collected on the 5 MinAtar games and averaged over 20 passes.
As shown before, whereas LaBER easily scales with larger $m$, it is not the case for DQN with large mini-batches, which are longer to run since the backward pass is the most costly operation.

\begin{table}
  \caption{CPU time comparison (in milliseconds)}
  \label{tab:time}
  \centering
\begin{tabular} { 
   m{5em} 
   m{7em}
   m{7em}
   m{7em}
   m{7em}
   }
 \hline
 DQN &  batch size $=B$ & batch size $=2B$ & batch size $=4B$ & batch size $=8B$ \\
 \hline
 Forward & 0.37 $\pm$ 0.02 & 0.52 $\pm$ 0.02 & 0.76 $\pm$ 0.03 & 1.3 $\pm$ 0.1 \\
 \hline
 Backward & 0.71 $\pm$ 0.02 & 0.90 $\pm$ 0.03 & 1.4 $\pm$ 0.1 & 2.6 $\pm$ 0.1 \\
 \hline\hline
 LaBER & $m=1$ & $m=2$ & $m=4$ & $m=8$ \\
 \hline
 Forward & 0.38 $\pm$ 0.02 & 0.52 $\pm$ 0.02 & 0.74 $\pm$ 0.03 & 1.2 $\pm$ 0.1 \\
 \hline
 Backward & 0.70 $\pm$ 0.03 & 0.71 $\pm$ 0.03 & 0.69 $\pm$ 0.03 & 0.70 $\pm$ 0.03 \\
\hline
\end{tabular}
\end{table}

\section{Detailed Results on MinAtar Games}
\label{app:minatar}

\begin{figure}
\begin{center}
\includegraphics[width=0.8\linewidth]{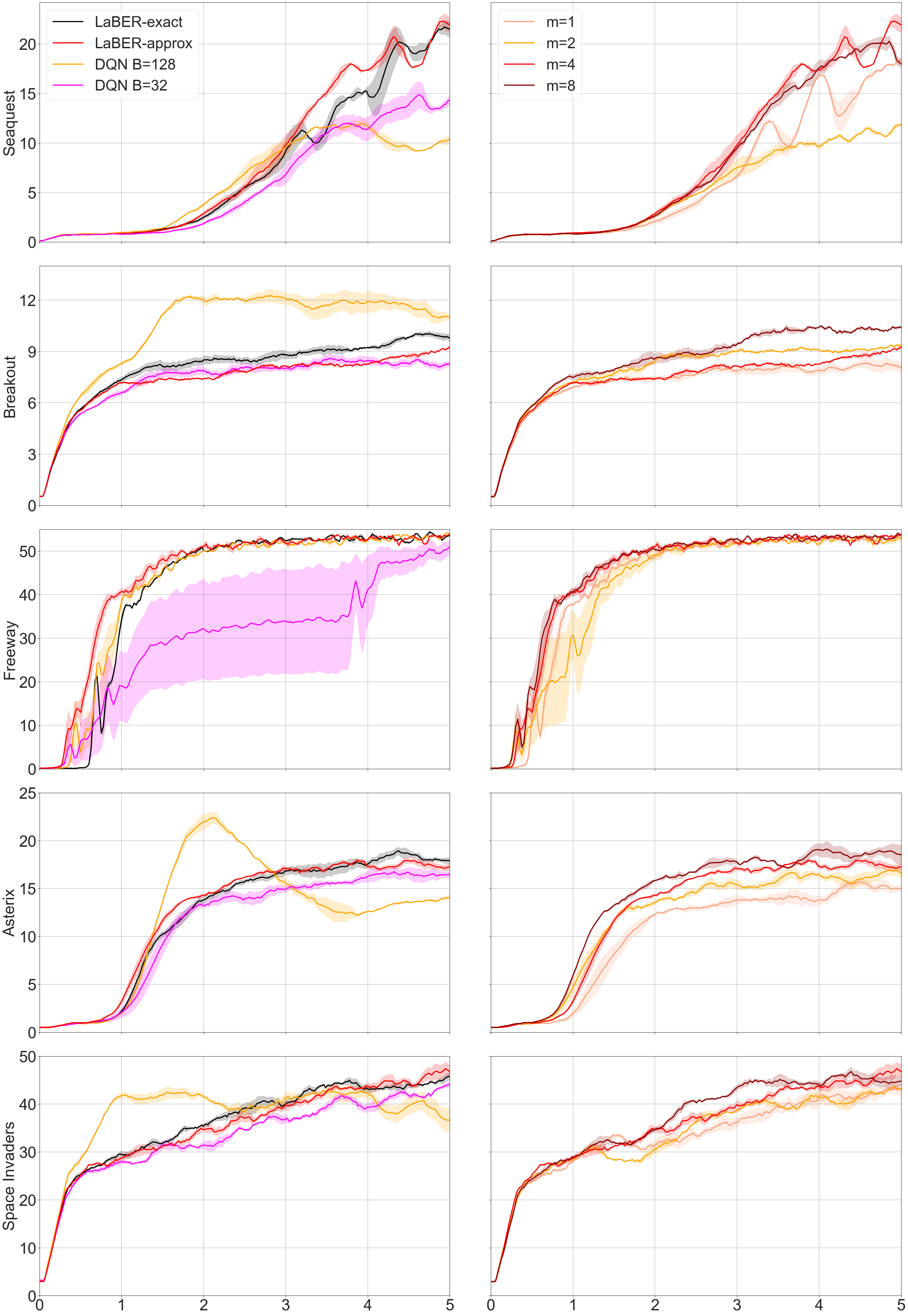}
\caption{On MinAtar games, the first column of this figure reports the performance of DQN with small and large mini-batches as well as LaBER with exact and approximate gradient norms. The second column reports the performance of LaBER with different values of $m$. The x-axis is the number of interaction steps in millions. The y-axis is the average sum of rewards gathered along each episode.}
\label{fig:minatar2_1}
\end{center}
\end{figure}

The first column of Figure \ref{fig:minatar2_1} reports the comparison of LaBER (LaBER-mean, $m=4$) with exact and approximate gradient norms and DQN with different mini-batch sizes on all MinAtar games. 
This column is the first column of Figure \ref{fig:minatar}. 

The second column of Figure \ref{fig:minatar2_1} reports the comparison of LaBER (LaBER-mean with approximate gradient norms) with different large batch sizes (different values of $m$). 
This column is the second column of Figure \ref{fig:minatar}. 

\begin{figure}
\begin{center}
\includegraphics[width=0.8\linewidth]{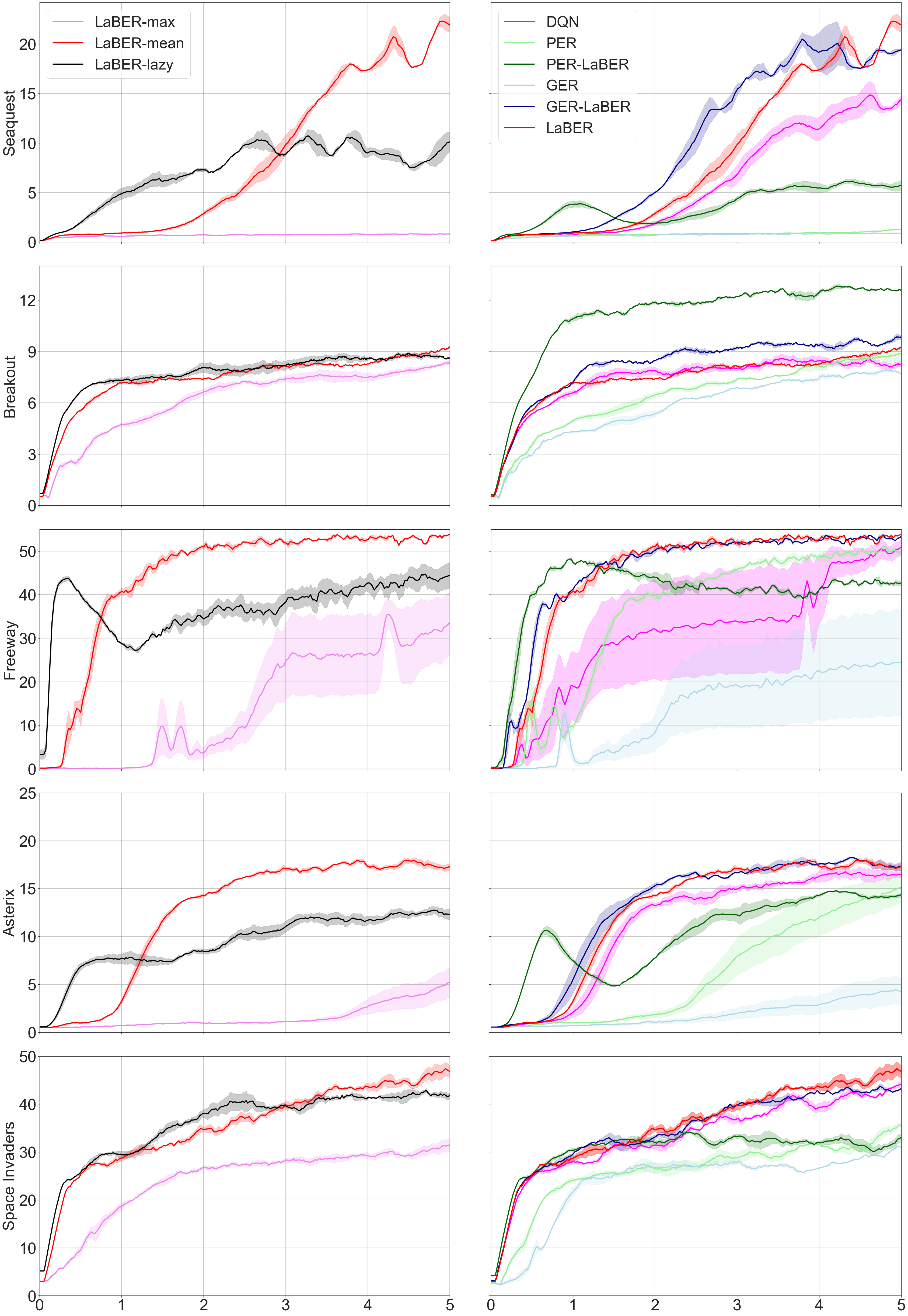}
\caption{On MinAtar games, the first column of this figure reports the performance of LaBER with different scaling for the gradient direction: LaBER-mean, LaBER-lazy and LaBER-max. The second column compares the performance of all studied agents: DQN, LaBER, PER, GER, PER-LaBER and GER-LaBER. The x-axis is the number of interaction steps in millions. The y-axis is the average sum of rewards gathered along each episode.}
\label{fig:minatar2_2}
\end{center}
\end{figure}

The first column of Figure \ref{fig:minatar2_2} reports the comparison of LaBER with different scaling for the descent direction (namely LaBER-mean, LaBER-lazy and LaBER-max) studied with $m=4$ and approximate gradient norms.
This column is the third column of Figure \ref{fig:minatar}. 

The second column of Figure \ref{fig:minatar2_2} reports the comparison of LaBER (LaBER-mean with approximate gradient norms and with $m=4$) and all other studied agents: DQN, PER, GER, PER-LaBER and GER-LaBER.
This column is the last column of Figure \ref{fig:minatar}.

\section{Detailed Results on Atari Games and PyBullet Environments}
\label{app:dopamine}

\begin{figure}
\begin{center}
\includegraphics[width=0.8\linewidth]{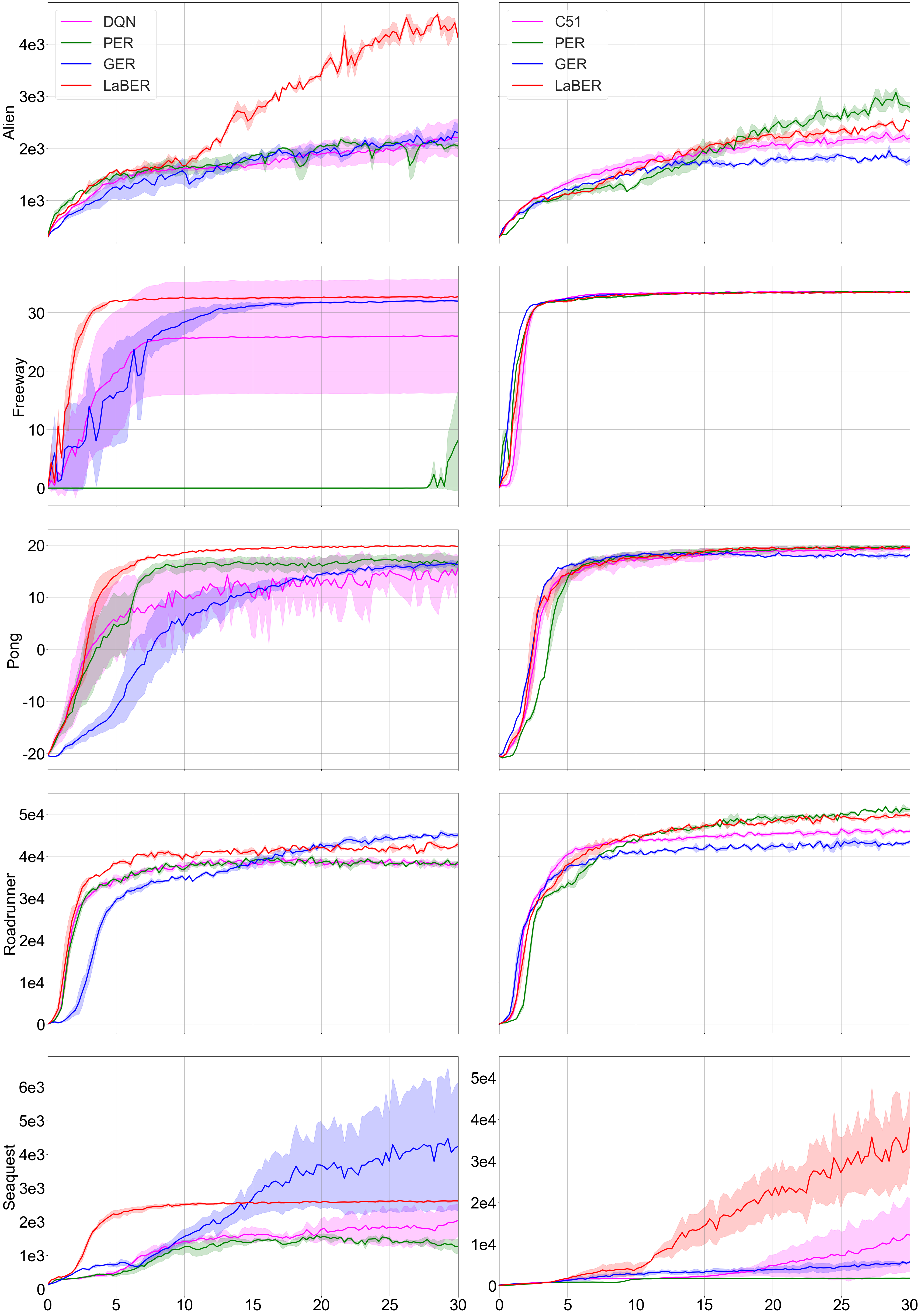}
\caption{On 5 Atari games, the first column of this figure reports the performance of PER, GER, LaBER (implemented over DQN) and DQN, whereas the second column reports the performance of PER, GER, LaBER (implemented over C51) and C51. The x-axis is the number of interaction steps in millions. The y-axis is the average Monte Carlo return computed every 250000 steps.}
\label{fig:dopamine2_1}
\end{center}
\end{figure}

The first column of Figure \ref{fig:dopamine2_1} reports the comparison of DQN, PER, GER and LaBER (LaBER-mean with surrogate gradient norms and with $m=4$) on 5 Atari games. 
This column is the first column of Figure \ref{fig:dopamine}. 

The second column of Figure \ref{fig:dopamine2_1} reports the comparison of C51, PER, GER and LaBER (LaBER-mean with surrogate gradient norms and with $m=4$) on the same 5 Atari games. 
This column is the second column of Figure \ref{fig:dopamine}. 
For PER, the value of the priority is the value of the loss, as proposed by \citet{Hessel2018Rainbow}.
On the contrary, LaBER uses the surrogate proposed in this work as priority.
As already stated at the end of Section \ref{sec:isavi}, this yields a very different value for prioritization. 
Appendix \ref{app:C51} derives the analytical expression of the surrogate gradient norms we use. 

\begin{figure}
\begin{center}
\includegraphics[width=0.8\linewidth]{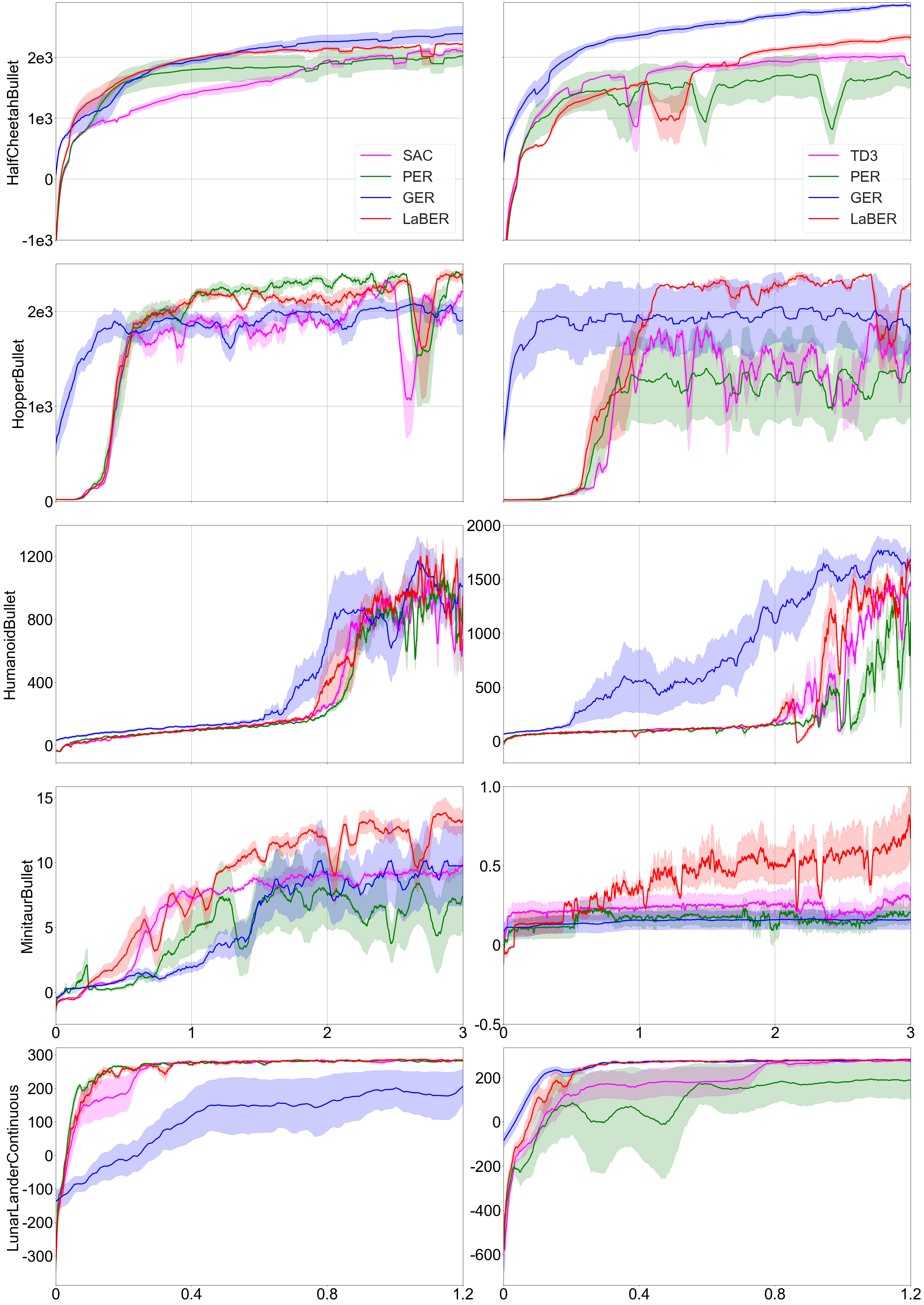}
\caption{On 4 PyBullet environments and LunarLander, the first column of this figure reports the performance of PER, GER, LaBER (implemented over SAC) and SAC, whereas the second column reports the performance of PER, GER, LaBER (implemented over TD3) and TD3. The x-axis is the number of interaction steps in millions. The y-axis is the average Monte Carlo return computed every 10000 steps.}
\label{fig:dopamine2_2}
\end{center}
\end{figure}

The first column of Figure \ref{fig:dopamine2_2} reports the comparison of SAC, PER, GER and LaBER (LaBER-mean with surrogate gradient norms and with $m=4$) on 4 PyBullet environments and LunarLander. 
This column is the third column of Figure \ref{fig:dopamine}. 

The second column of Figure \ref{fig:dopamine2_2} reports the comparison of TD3, PER, GER and LaBER (LaBER-mean with surrogate gradient norms and with $m=4$) on 4 PyBullet environments and LunarLander. 
This column is the last column of Figure \ref{fig:dopamine}.

Since two critic networks are used in TD3 or SAC, two per-sample gradient norms are available, and the appropriate way of dealing with this two quantities in PER, GER or LaBER deserves some attention.
In our GitHub repository, we dedicate a didactic folder aiming at clarifying this issue. See: \url{https://github.com/sureli/laber}.

Whereas \citet{fujimoto2020equivalence} implement PER over TD3 by using the maximum of the two TD errors given by the two critic networks as priorities, we consider, according to what we derived in our work, that each network must be learnt with dedicated samples. 
Indeed, each network has its own optimal sampling distribution $p^*_i \propto \| \nabla_\theta \ell(Q_\theta(x_i), y_i) \|_2$ over the $N$ items of the replay buffer.
Hence, for PER and GER, two lists of priorities are maintained, and the SGD step of each network uses transitions sampled according to the distribution defined by the corresponding list of priorities.
For LaBER, once the large batch is uniformly sampled, the up-to-date (exact or approximate) per-sample gradient norms are computed separately by each critic, and down-sampled separately. 
The samples used for the SGD update are not the same for the two critics.

\end{document}